\PassOptionsToPackage{numbers,sort&compress}{natbib}
\documentclass[letterpaper,11pt]{article} 
\usepackage[margin=1in]{geometry}
\usepackage{graphicx}
\usepackage{hyperref}     
\usepackage{natbib}       
\usepackage{authblk}      

\usepackage[margin=1in]{geometry}
\usepackage{microtype}
\usepackage{lmodern}
\usepackage[T1]{fontenc}
\usepackage[utf8]{inputenc}

\usepackage{amsmath,amssymb,amsthm,mathtools}

\usepackage{graphicx}
\usepackage{booktabs}
\usepackage{multirow}
\usepackage{siunitx}
\usepackage{subcaption}
\usepackage{float}

\usepackage{xcolor}
\usepackage{enumitem}

\usepackage{hyperref}
\hypersetup{
  colorlinks=true,
  linkcolor=blue,
  citecolor=blue,
  urlcolor=blue
}

\usepackage{algorithm}
\usepackage{algorithmic}

\usepackage[numbers,sort&compress]{natbib}

\usepackage[capitalise,noabbrev]{cleveref}

\usepackage[font=small,labelfont=bf]{caption}

\usepackage{url}

\usepackage{makecell}

\usepackage{placeins}

\usepackage{authblk}

\usepackage[T1]{fontenc}
\usepackage[utf8]{inputenc} 
\DeclareUnicodeCharacter{2248}{\ensuremath{\approx}}

\usepackage{threeparttable}
\usepackage{booktabs}     
\usepackage{graphicx}     

\title{StreetMath: Study of LLMs’ Approximation Behaviors}

\author[1]{Chiung-Yi Tseng}
\author[2]{Somshubhra Roy}
\author[3]{Maisha Thasin}
\author[4]{Danyang Zhang}
\author[5]{Blessing Effiong}

\affil[1]{LuxMuse AI \\ \texttt{\href{mailto:ctseng@luxmuse.ai}{ctseng@luxmuse.ai}}}
\affil[2]{Department of Electrical and Computer Engineering, North Carolina State University \\ \texttt{\href{mailto:sroy22@ncsu.edu}{sroy22@ncsu.edu}}}
\affil[3]{Department of Mathematics, University of Waterloo \\ \texttt{\href{mailto:thasin.maisha@gmail.com}{thasin.maisha@gmail.com}}}
\affil[4]{Vokram Group. \\ \texttt{\href{mailto:danyang@vokram.com}{danyang@vokram.com}}}
\affil[5]{Department of Computer Science, Saint Louis University \\ \texttt{\href{mailto:Blessing.effiong@slu.edu}{Blessing.effiong@slu.edu}}}

\date{} 

\usepackage{bibentry}

\begin{document}

\maketitle
\begin{abstract}
There is a substantial body of literature examining the mathematical reasoning capabilities of large language models (LLMs), particularly their performance on precise arithmetic operations in autoregressive architectures. However, their ability to perform approximate reasoning in informal, fast-paced mathematical operations has received far less attention, especially among non-autoregressive decoder models. Our work addresses this gap by introducing StreetMath, a benchmark designed to evaluate models’ approximation abilities under real-world approximation scenarios. We conduct extensive evaluations across different LLM architectures: Qwen3-4B-Instruct-2507, Qwen3-4B-Thinking-2507, Dream-v0-Instruct-7B, Falcon-Mamba-7B-Instruct, and Mamba-GPT-3B. Furthermore, we apply mechanistic interpretability techniques to probe their internal computational states. Our analysis reveals that LLMs generally attempt to compute exact values or invoke external tools even in tasks that call for approximation. Moreover, while models sometimes reach the correct answer in early layers or steps, they still consume more tokens when solving approximation tasks. Additional experiments indicate that exact and approximate arithmetic operations rely on largely separate neural components. Drawing upon research on cognitive psychology, we argue that LLMs do not exhibit cognitive miserliness in the same way humans do in street math settings. We open source our work \url{https://github.com/ctseng777/StreetMath}
\end{abstract}

\section{Introduction}


Human mathematical reasoning flexibly alternates between exact calculation and rough estimation, depending on context. This adaptability—often described as "cognitive miserliness"\cite{kahneman2011thinking}—allows people to conserve effort by using approximations when precision is unnecessary. According to Kahneman's dual-process theory, humans preferentially rely on System 1 (fast, intuitive) thinking for everyday approximate calculations—what we call \emph{street math}—the quick mental calculations people make in everyday life, such as estimating the total cost of groceries or computing a restaurant tip (e.g., leaving a 20\% tip on a \$61 bill—roughly 20\% of \$60 ≈ \$12, which is much easier to calculate). This reflects the broader concept of cognitive miserliness, as the adaptive tendency to minimize mental effort by employing shortcuts and approximations when full precision is unnecessary \cite{fiske1991social}. Street math exemplifies the context where System 1 dominates: quick estimates suffice, and the cognitive cost of engaging System 2 (slow, effortful) reasoning is unwarranted. This principle also highlights fundamental capacity limitations: cognitive processing requires effort, which humans are motivated to conserve by using "good enough" strategies when circumstances permit. Our findings reveal that large language models (LLMs), in contrast, tend to bypass this adaptive flexibility. Instead of switching to easier approximation when appropriate, they engage in effortful, exact computation—even when rapid estimation would be more efficient—paralleling a departure from human-like cognitive efficiency. Recent interpretability studies have uncovered Fourier-like computation circuits~\cite{zhou2024pretrained} and attention heads dedicated to mathematical processing~\cite{yu2024interpreting}. Yet it remains unclear whether these models exhibit the same context-sensitive flexibility as humans, or whether their reasoning is rigidly tied to exact solutions.

In this work, we introduce the \emph{StreetMath} dataset, a curated collection of 1000 approximation problems drawn from everyday street math scenarios. Using this benchmark, we systematically evaluate diverse model classes, including autoregressive decoder architectures (Qwen3-4B-Instruct-2507~\cite{qwen3technicalreport}, Qwen3-4B-Thinking-2507), state-space models (Falcon-Mamba-7B~\cite{zuo2024falconmambacompetitiveattentionfree}, Mamba-GPT-3B~\cite{mambagpt3b_huggingface}), and diffusion-based language models (Dream-v0-Instruct-7B~\cite{ye2025dream}). Our experiments reveal a consistent bias across all architectures: models overwhelmingly favor exact computation, even in contexts where rough estimation would suffice. Most importantly, some models achieve better approximation scores only at the cost of increased computation (tokens), which runs counter to humans’ cognitive miserliness. To better understand this limitation, we examine models’ rounding behavior, a fundamental operation for approximation in the street math setting. We apply linear probing to compare internal representations, finding that models’ approximation on single numbers resembles human behavior: they often round numbers toward 5 or 10. In addition, models perform well at digit-level detection but struggle to generalize to word-based numbers~\cite{levy2024encode}. 

We further investigate the neural underpinnings of these behaviors. By pruning the neurons involved in exact arithmetic~\cite{christ2025math}, we uncover a surprising dynamic: removing math-specific parameters can actually improve performance on approximation tasks. This suggests that rigid, precision-oriented circuits may actively hinder flexible estimation. Additional probing into the entropy and effective ranks of intermediate layers~\cite{skean_layer_2025} reveals similar distributions and dimensionalities between exact arithmetic operations and approximation. These findings imply that approximation does not reduce computational cost—contrary to how humans use approximation to simplify computation.

Together, these findings suggest that while LLMs have developed specialized pathways for arithmetic, they lack the human-like adaptability required for context-sensitive street math. Although LLMs are capable of approximating single numbers, they do not leverage this ability \emph{during} the process of solving street math questions; instead, they approximate only after calculating exact answers. We conclude that LLMs do not reason about approximation questions in the same way humans do. The training corpora likely introduce this universal gap across model architectures and sizes.

\section{StreetMath Dataset \& Evaluations}
\label{benchmark}
We release 1,000 multiple-choice math reasoning problems under street math settings, covering five major topics, each with several subtopics: basket sum (sum of shopping items), discounts (buy-$n$-get-$m$-free, threshold discounts such as ``\$X off if you spend \$Y", percentage discounts), taxes (tax before discount and tax after discount applied), units (calculating cost based on per-pound or per-kilogram prices), and tips (\% on spend). Each question offers four answer options, designed to distinguish different levels of approximation capability: exact calculation, good approximation (within 20\% relative error of the exact answer), mildly off (between 60\% and 90\% relative error), and way off (greater than 150\% relative error). Details of the benchmark is elaborated in \ref{App_BenchmarkingResults}. The benchmark not only evaluates final answers but also examines intermediate numerical evidence and the chain-of-thought (CoT) reasoning process. Any traces of exact computation or tool usage are flagged as exact math. To assess whether models exhibit cognitive miserliness, we use token count as a proxy for reasoning efficiency.

We evaluate a range of model architectures including \textbf{ autoregressive decoder}, \textbf{ state-space }and \textbf{language diffusion models} across different reasoning styles (CoT vs. non-CoT) and parameter sizes (3B, 4B, 7B). The models include Qwen3-4B-Instruct-2507, Qwen3-4B-Thinking-2507, Dream-v0-Instruct-7B, Falcon-Mamba-7B-Instruct, and mamba-GPT-3B, with experiemnt setup details in \Ref{App_Experiment_setup}. We carefully adapt system and user prompts to each architecture to ensure fair comparisons. As shown in Table\ref{table:benchmark} and  \ref{table:benchmark_by_topic}, LLMs across all architectures predominantly compute exact answers even when model prompt explicitly asks for approximation. When they do produce approximated answers, they typically first compute the exact value and then round it. Notably, Qwen3-4B-Thinking-2507 shows better approximation performance than Qwen3-4B-Instruct-2507, but this improvement comes at the cost of higher token usage (228 vs. 125 tokens on average) and increased deviations contrary to human cognitive miserliness. State-space models achieve similar approximation performance to Qwen3-4B-Instruct-2507 with fewer tokens but greater deviations. Dream-v0-Instruct-7B consistently produces exact answers with perfect accuracy. We leave it to future work to investigate whether adjusting the steps and temperatures of Dream-v0-Instruct-7B can improve its approximation performance.

Overall, our findings indicate that LLMs tend to rely on exact arithmetic even in approximation settings, showing behavior opposite to human-like cognitive miserliness.

\begin{table*}[t]
\centering
  \centering
  \begin{threeparttable}

  \begin{tabular}{lrrrrrrr}
  \toprule
  Model & A & E & M & W & Uncategorized & Tool calls & Avg tokens \\
  \midrule
  Qwen3-4B-Instruct-2507   & 445 & 514 & 40  & 1   & 0   & 1000 & 125 \\
  Qwen-4B-Thinking-2507    & 151 & 637 & 197 & 15  & 0   & 0    & 228 \\
  Dream-v0-Instruct-7B     & 0   & 1000& 0   & 0   & 0   & 0    & 263 \\
  Falcon-Mamba-7B-Instruct          & 177 & 469 & 131 & 22  & 201 & 0    & 131  \\
  Mamba-GPT-3B             & 174 & 459 & 166 & 198 & 3   & 0    & 86  \\ [1ex]
  \bottomrule
  \end{tabular}
  \begin{tablenotes}
  \item Abbreviations: A = Good approximation, E = Exact Math, M = Mildly off, W = Way off
  \end{tablenotes}
  \end{threeparttable}
  \caption{Overall judgement counts by model with tool calls and average tokens (rounded).}
  \label{table:benchmark}
\end{table*}
\begin{table}[t]
  \centering
  \resizebox{\textwidth}{!}{
  \begin{tabular}{l l r r r r r r}
  \toprule
  Model & Topic & Good approx & Exact math & Mildly off & Way off & Uncategorized & N \\
  \midrule
  Qwen3-4B-\\Instruct-2507 & basket\_sum & 86 & 154 & 1 & 0 & 0 & 241 \\
   & discounts & 15 & 220 & 7 & 0 & 0 & 242 \\
   & taxes & 40 & 132 & 1 & 0 & 0 & 173 \\
   & units & 22 & 150 & 0 & 0 & 0 & 172 \\
   & tips & 22 & 150 & 0 & 0 & 0 & 172 \\
  Qwen-4B-\\Thinking-2507 & basket\_sum & 46 & 104 & 55 & 36 & 0 & 241 \\
   & discounts & 80 & 61 & 51 & 50 & 0 & 242 \\
   & taxes & 40 & 45 & 46 & 42 & 0 & 173 \\
   & units & 35 & 84 & 22 & 31 & 0 & 172 \\
   & tips & 28 & 68 & 40 & 36 & 0 & 172 \\
  Dream-v0-\\Instruct-7B & basket\_sum & 0 & 241 & 0 & 0 & 0 & 241 \\
   & discounts & 0 & 242 & 0 & 0 & 0 & 242 \\
   & taxes & 0 & 173 & 0 & 0 & 0 & 173 \\
   & units & 0 & 172 & 0 & 0 & 0 & 172 \\
  & tips & 0 & 172 & 0 & 0 & 0 & 172 \\
  Falcon-Mamba-7B & basket\_sum & 47 & 106 & 43 & 0 & 45 & 241 \\
   & discounts & 50 & 108 & 61 & 5 & 18 & 242 \\
   & taxes & 38 & 63 & 47 & 0 & 25 & 173 \\
   & units & 8 & 94 & 7 & 14 & 49 & 172 \\
   & tips & 11 & 77 & 4 & 0 & 80 & 172 \\
  Mamba-GPT-3B & basket\_sum & 51 & 97 & 46 & 47 & 0 & 241 \\
   & discounts & 43 & 111 & 35 & 53 & 0 & 242 \\
   & taxes & 29 & 59 & 39 & 43 & 3 & 173 \\
   & units & 32 & 78 & 31 & 31 & 0 & 172 \\
   & tips & 19 & 114 & 15 & 24 & 0 & 172 \\
  \bottomrule
  \end{tabular}
  }
  \caption{Benchmark results: Counts by topic for all models.}
  \label{table:benchmark_by_topic}
\end{table}

\FloatBarrier

\section{Linear Probe on Rounding Behaviors}
\label{linear_probe}

\begin{figure*}[!htb]
\centering
 \begin{subfigure}{0.8\linewidth}
        \centering
        \includegraphics[width=\linewidth]{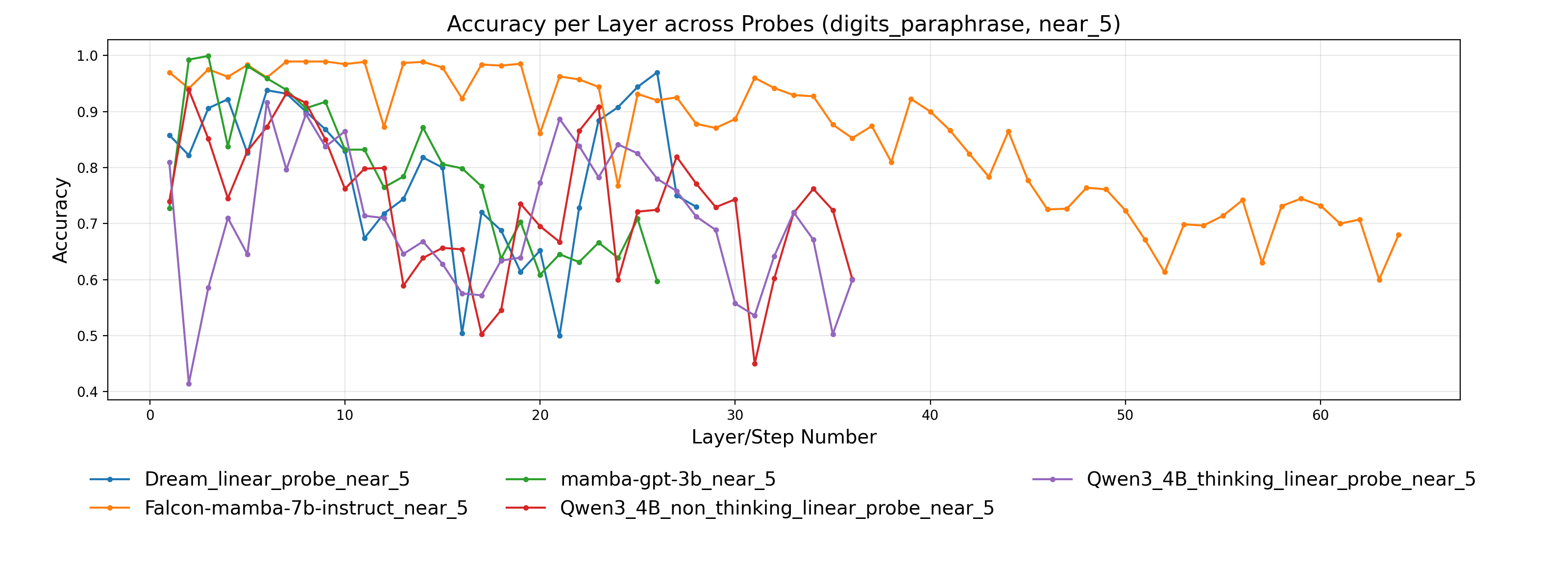}
        \subcaption{Digits paraphrase (near=5)}
        \label{fig:digits_paraphrase_near_5}
    \end{subfigure}
 
    \begin{subfigure}{0.8\linewidth}
        \centering
        \includegraphics[width=\linewidth]{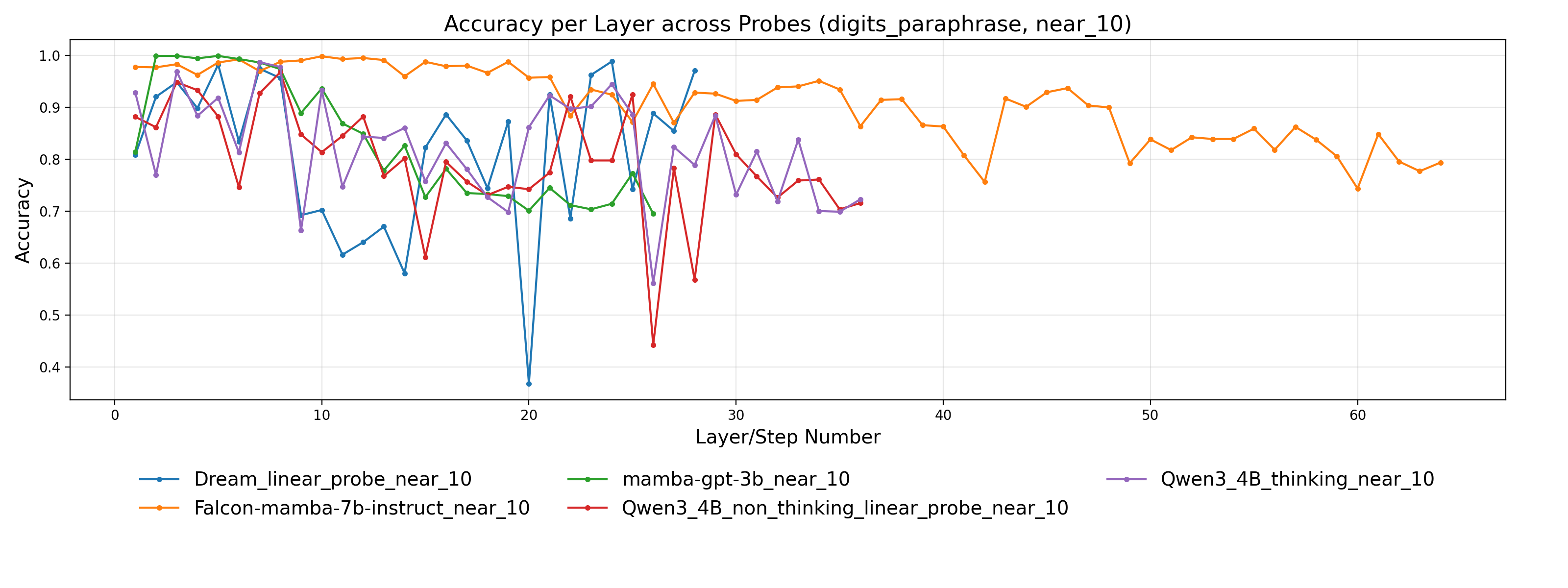}
        \subcaption{Digits paraphrase (near=10)}
        \label{fig:digits_paraphrase_near_10}
    \end{subfigure}
    
    \begin{subfigure}{0.8\linewidth}
        \centering
        \includegraphics[width=\linewidth]{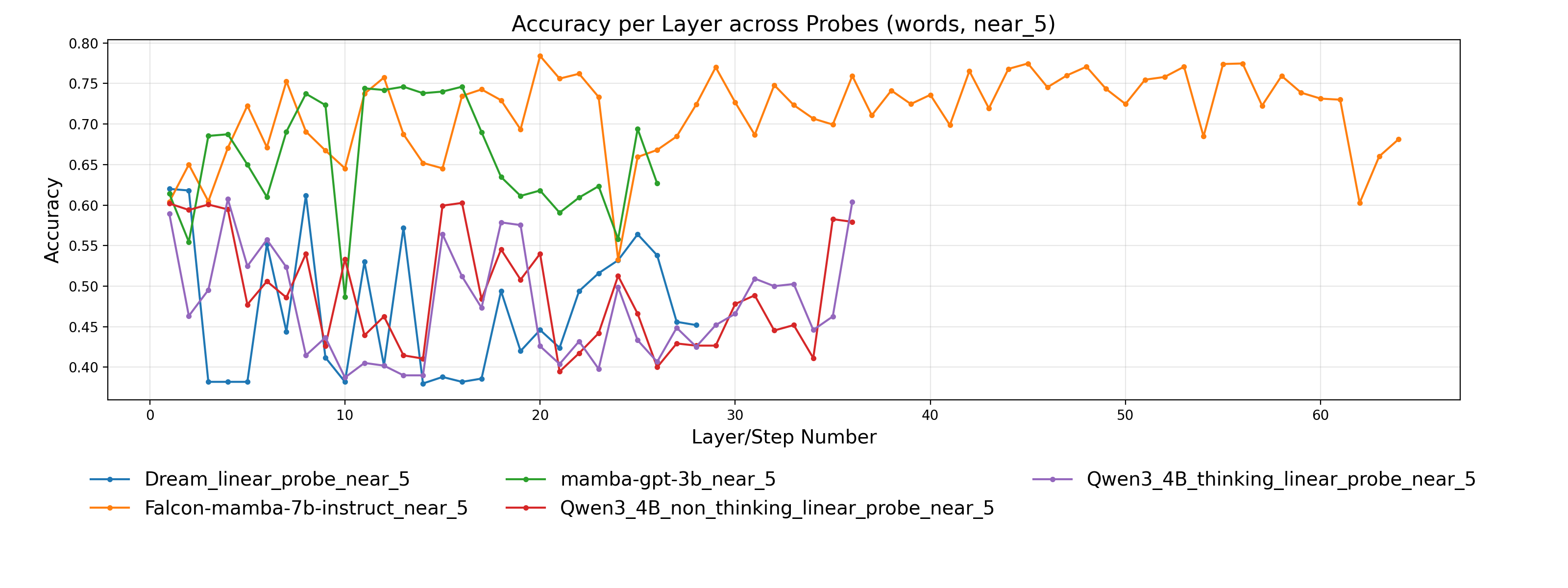}
        \subcaption{Words (near=5)}
        \label{fig:words_near_5}
    \end{subfigure}
    
    \begin{subfigure}{0.8\linewidth}
        \centering
        \includegraphics[width=\linewidth]{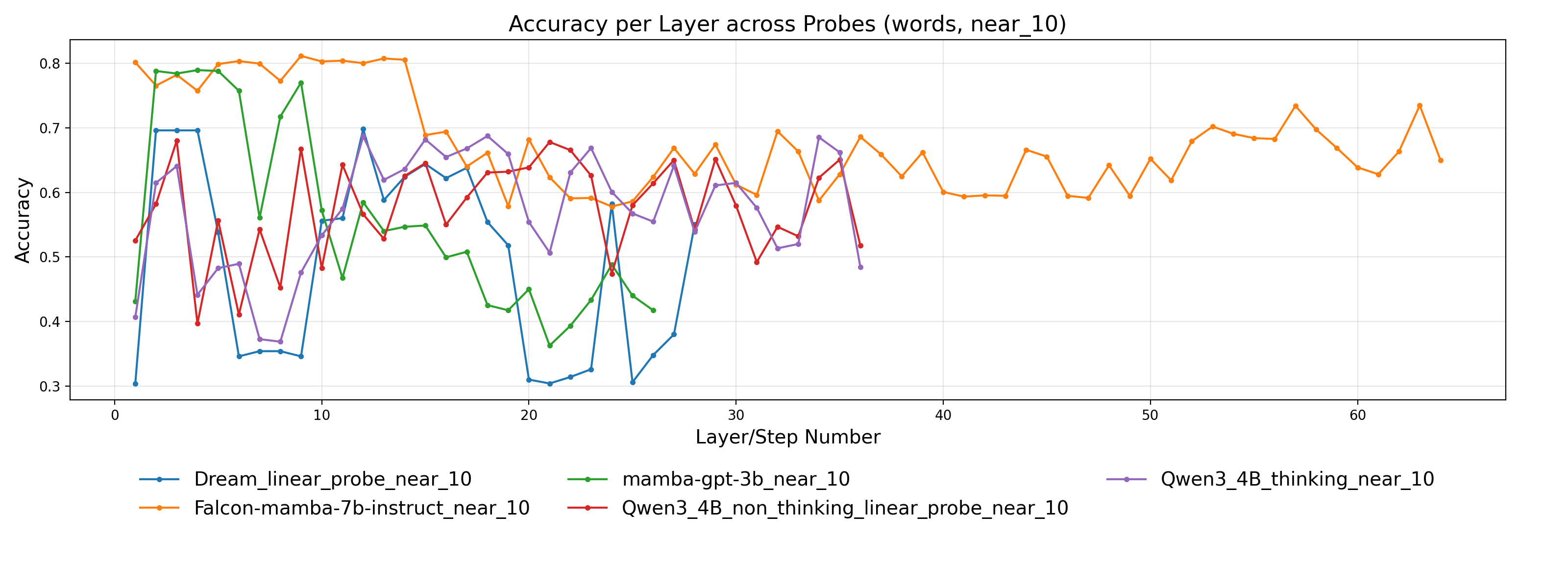}
        \subcaption{Words (near=10)}
        \label{fig:words_near_10}
    \end{subfigure}
    
    \caption{Accuracy per layer across models for digits paraphrase and words tasks with near parameters 5 and 10.}
    \label{fig:linear_probe_grid}
\end{figure*}

We investigate whether models encode numerical topology similar to human cognitive distance effects~\cite{dehaene2011number,moyer1967time} by training linear probes~\cite{alain2016understanding,hewitt2019structural} to detect nearness to multiples of 5 and 10~\cite{debrauwer2006five}, defining proximity as exactly one integer away from the nearest multiple (e.g., 21 is near-10; 22 is not). Using simple templates to extract hidden-state representations, we evaluate five StreetMath models on digit-based (“Here is 23.”) and word-based (“Consider the number twenty three.”) inputs, analyzing  
(i) layer-wise accuracy,  
(ii) best-layer errors across distances 0, 1, 2+. The experiment setup is elaborated in \ref{App_LinearProbe}, and results are shown in Figure \ref{fig:linear_probe_grid} and Table \ref{tab:near5} to Table \ref{tab:near10_words}.  

Digit tasks show early emergence~\cite{teerapittayanon2016branchynet} where state-space models lead: Mamba-GPT-3B reaches 99.9\% and Falcon-Mamba-7B reaches 98\%, with best layers in early–middle positions (shortcut-friendly; supports early stopping), whereas Dream-v0-Instruct-7B peaks late (26th Near-5, 24th Near-10), consistent with diffusion vs. autoregressive/state-space differences. Distance-1 cases (e.g., 9, 11, 14, 16) are hardest, reflecting digit encoding~\cite{levy2025language} and calibration biases~\cite{lovering2024language}. Word tasks underperform across architectures, evidencing surface-form encoding and limited numerical abstraction~\cite{mccoy2019right,belinkov2019analysis,goldberg2016primer}, likely due to tokenization, pretraining bias toward digits, and separable digit/word representational clusters.

\begin{table*}[!htb]
  \centering

  \begin{tabular}{|>{\raggedright\arraybackslash}
  p{0.3\textwidth}|c|c|c|c|c|}
  \hline
  \textbf{Model} & \textbf{Peak Acc} & \textbf{Best
  Layer} & \textbf{Err (0)} & \textbf{Err (1)} &
  \textbf{Err (2)} \\
  \hline
  \makecell[tl]{Qwen3-4B-Instruct} & 0.939 & 2 & 0.4\%
  & 5.5\% & 9.4\% \\
  \hline
  \makecell[tl]{Qwen3-4B-Thinking} & 0.917 & 6 & 7.2\% &
  14.6\% & 2.5\% \\
  \hline
  \makecell[tl]{Dream-7B} & 0.970 & 26 & 4.2\% & 4.8\% & 0.5\% \\
  \hline
  \makecell[tl]{Falcon-Mamba-7B\\-Instruct} & 0.989 & 7 &
  0.7\% & 0.6\% & 1.7\% \\
  \hline
  \makecell[tl]{Mamba-GPT-3B} & 0.999 & 3 & 0.4\% & 0.0\%
  & 0.0\% \\
  \hline
  \end{tabular}
    \caption{Comprehensive Near-5 Digit Analysis: Performance and
  Error Patterns at the best layer. Acc = Accuracy; Err =
  Error rate}
  \label{tab:near5}
  \end{table*}
\begin{table*}[!htb]
  \centering

  \begin{tabular}{|>{\raggedright\arraybackslash}
  p{0.3\textwidth}|c|c|c|c|c|}
  \hline
  \textbf{Model} & \textbf{Peak Acc} & \textbf{Best
  Layer} & \textbf{Err (0)} & \textbf{Err (1)} &
  \textbf{Err (2)} \\
  \hline
  \makecell[tl]{Qwen3-4B-Instruct} & 0.603 & 16 &
  7.0\% & 4.0\% & 94.3\% \\
  \hline
  \makecell[tl]{Qwen3-4B-Thinking} & 0.607 & 4 & 0.4\% &
  0.6\% & 100.0\% \\
  \hline
  Dream-7B & 0.620 & 1 & 0.0\% & 0.0\% & 99.5\% \\
  \hline
  \makecell[tl]{Falcon-Mamba-7B\\-Instruct} & 0.784 & 20 &
  4.2\% & 2.7\% & 50.5\% \\
  \hline
  \makecell[tl]{Mamba-GPT-3B} & 0.746 & 13 & 2.1\% & 0.0\%
  & 64.2\% \\
  \hline
  \end{tabular}
  \caption{Comprehensive Near-5 (Words) Analysis:
  Performance and Error Patterns at the best layer. Acc =
  Accuracy; Err = Error rate}
  \label{tab:near5_words}
  \end{table*}
\begin{table*}[!htb]
  \centering
  \resizebox{\textwidth}{!}{%
  \begin{tabular}{|>{\raggedright\arraybackslash}
  p{0.25\textwidth}|c|c|c|c|c|c|c|}
  \hline
  \textbf{Model} & \textbf{Peak Acc} & \textbf{Best Layer} &
  \textbf{Err (0)} & \textbf{Err (1)} & \textbf{Err (2)}
  & \textbf{Err (3)} & \textbf{Err (4+)} \\
  \hline
  \makecell[tl]{Qwen3-4B-Instruct} & 0.967 & 8 & 4\% &
  12\% & 1\% & 1\% & 0\% \\
  \hline
  \makecell[tl]{Qwen3-4B-Thinking} & 0.987 & 7 & 1\% & 3\%
  & 3\% & 0\% & 1\% \\
  \hline
  \makecell[tl]{Dream-7B} & 0.988 & 24 & 2\% & 5\% & 0\% &
  0\% & 0\% \\
  \hline
  \makecell[tl]{Falcon-Mamba-7B\\-Instruct} & 0.998 & 10 &
  1\% & 0\% & 1\% & 0\% & 0\% \\
  \hline
  \makecell[tl]{Mamba-GPT-3B} & 0.999 & 2 & 0\% & 0\% & 0\%
  & 0\% & 0\% \\
  \hline
  \end{tabular}
  }
  \caption{Comprehensive Near-10 Analysis: Performance
  and Error Patterns at the Best Layer}
  \label{tab:near10}
  \end{table*}
\begin{table*}[!htb]
  \centering
  \resizebox{\textwidth}{!}{%
  \begin{tabular}{|>{\raggedright\arraybackslash}
  p{0.25\textwidth}|c|c|c|c|c|c|c|}
  \hline
  \textbf{Model} & \textbf{Peak Acc} & \textbf{Best
  Layer} &
  \textbf{Err (0)} & \textbf{Err (1)} & \textbf{Err (2)}
  & \textbf{Err (3)} & \textbf{Err (4+)} \\
  \hline
  \makecell[tl]{Qwen3-4B-Instruct} & 0.680 & 3 & 96\% &
  98\% & 3\% & 4\% & 3\% \\
  \hline
  \makecell[tl]{Qwen3-4B-Thinking} & 0.687 & 18 & 97\% &
  96\% & 4\% & 2\% & 2\% \\
  \hline
  \makecell[tl]{Dream-7B} & 0.698 & 12 & 98\% & 100\% &
  0\% &
  0\% & 0\% \\
  \hline
  \makecell[tl]{Falcon-Mamba-7B\\-Instruct} & 0.811 & 9 &
  67\% & 58\% & 0\% & 0\% & 0\% \\
  \hline
  \makecell[tl]{Mamba-GPT-3B} & 0.789 & 4 & 74\% & 57\%
  & 2\%
  & 5\% & 2\% \\
  \hline
  \end{tabular}
  }
  \caption{Comprehensive Near-10 (Words) Analysis:
  Performance
  and Error Patterns at the Best Layer}
  \label{tab:near10_words}
  \end{table*}

\captionsetup[subfigure]{skip=-1pt} 

\FloatBarrier

\section{Causal Studies}
\label{MathNeurons}
To isolate parameters tied to exact arithmetic \cite{christ_math_2025,rai_practical_2025}, We adapt the \textbf{MathNeuro} codebase to study pruning and scaling in instruction-tuned LMs, with experiment details in \ref{App_Causal}. For each calibration corpus (a CSV with \textit{instruction} and \textit{response} columns), we estimate parameter importance by registering forward hooks on all \texttt{Linear} layers and accumulating mean activation magnitudes weighted by the corresponding weight magnitudes over 200 calibration samples. We then construct a keep-mask that retains the top $p\%$ of parameters, where $p \in \{0.01\%, 0.1\%, 0.5\%, 1\%, 2.5\%, 5\%, 10\%, 25\%, 50\%\}$. 

\begin{figure*}
    \centering

    \begin{subfigure}{0.49\textwidth}
        \centering
        \includegraphics[width=\linewidth]{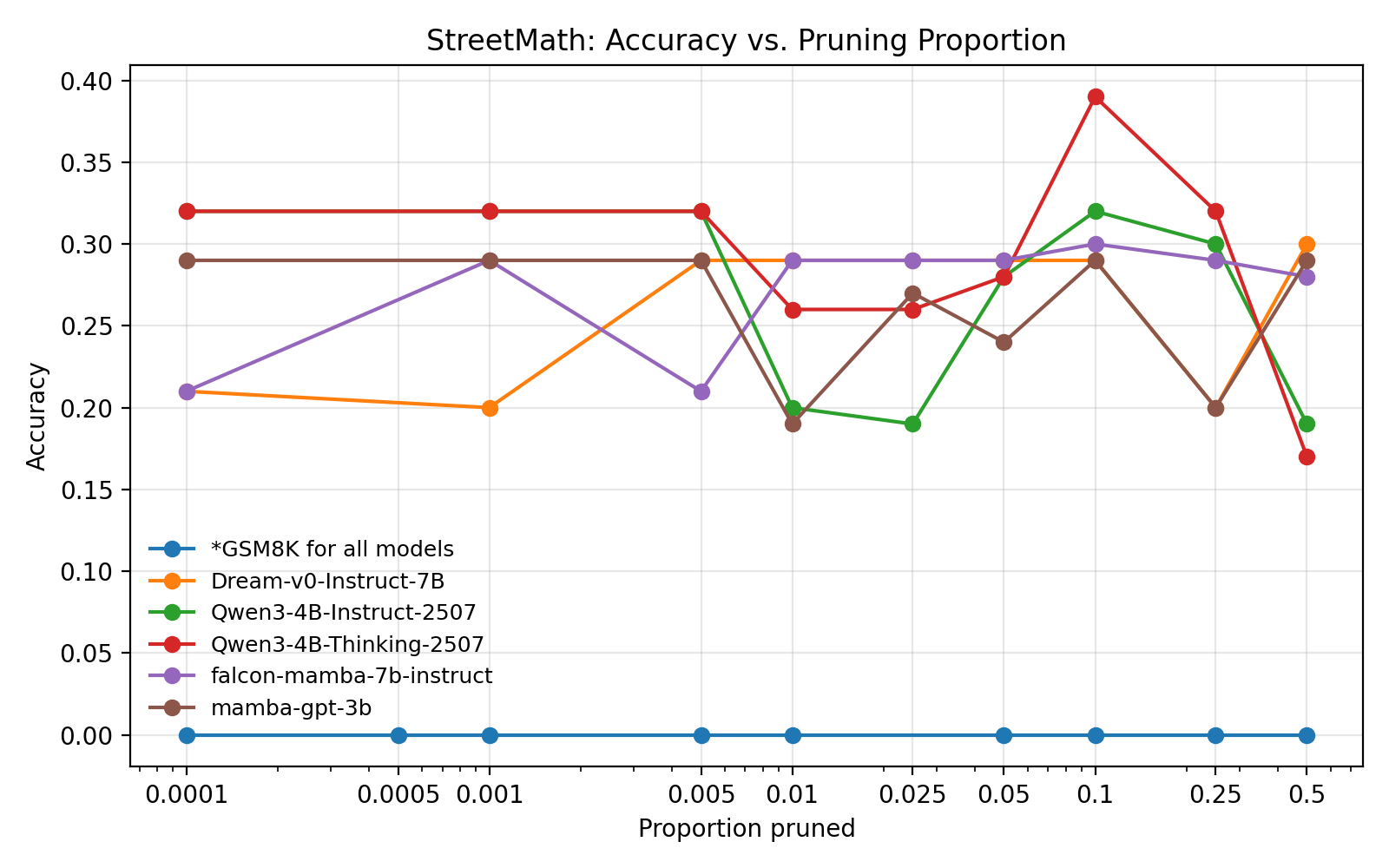}
        \subcaption{Overall accuracy}
        \label{fig:place_holder}
    \end{subfigure}
    \hfill
    \begin{subfigure}{0.49\textwidth}
        \centering
        \includegraphics[width=\linewidth]{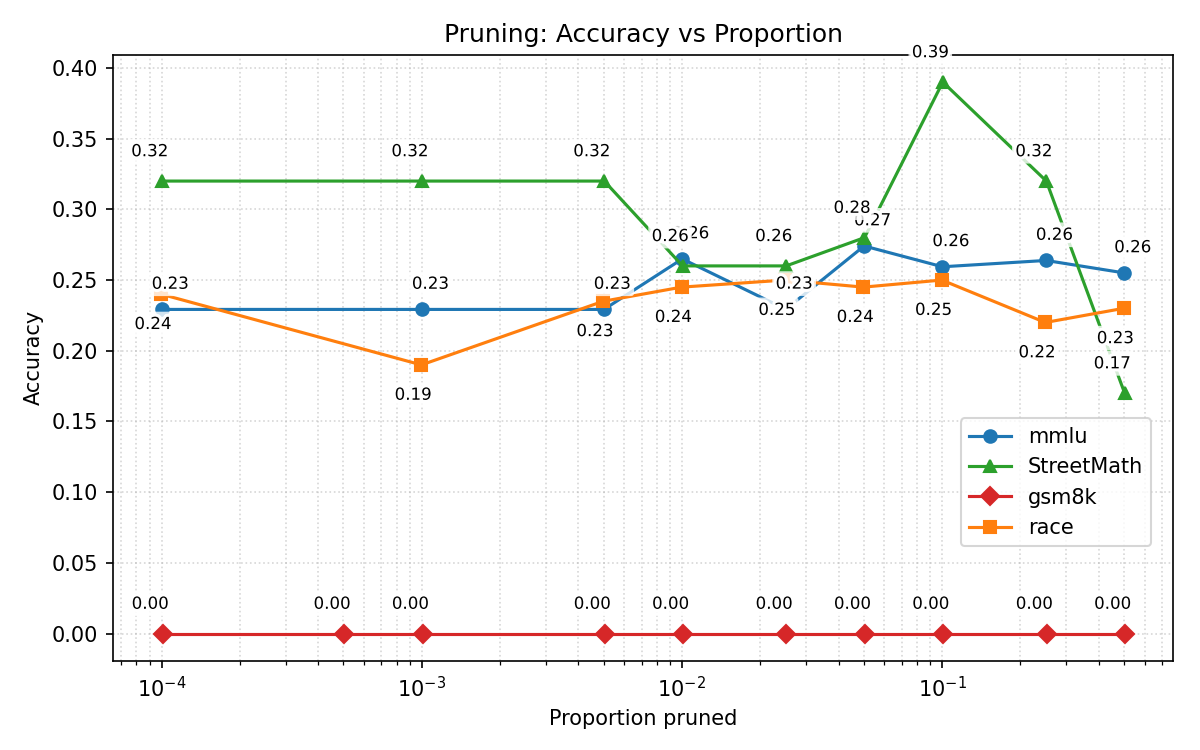}
        \subcaption{Pruning accuracy on Qwen3-4B-Thinking-2507}
        \label{fig:place_holder}
    \end{subfigure}

    \vspace{0.5em}

    \begin{subfigure}{0.49\textwidth}
        \centering
        \includegraphics[width=\linewidth]{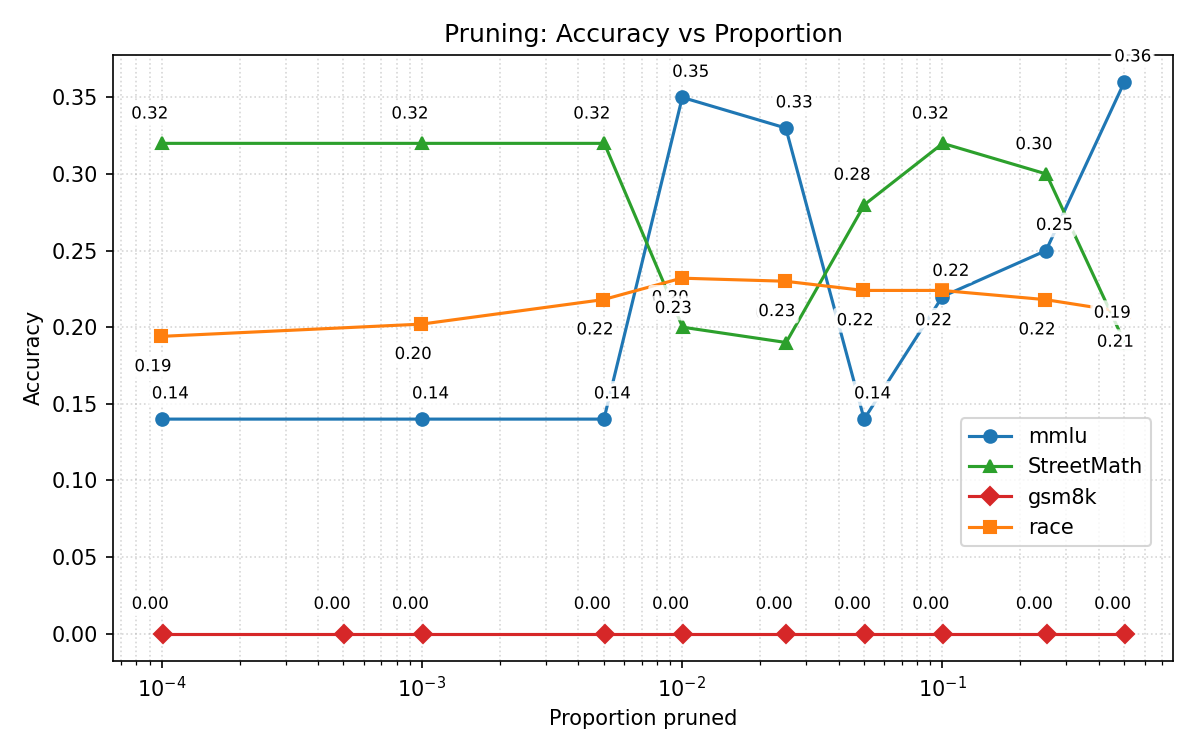}
        \subcaption{Pruning accuracy on Qwen3-4B-Instruct-2507}
        \label{fig:place_holder}
    \end{subfigure}
    \hfill
    \begin{subfigure}{0.49\textwidth}
        \centering
        \includegraphics[width=\linewidth]{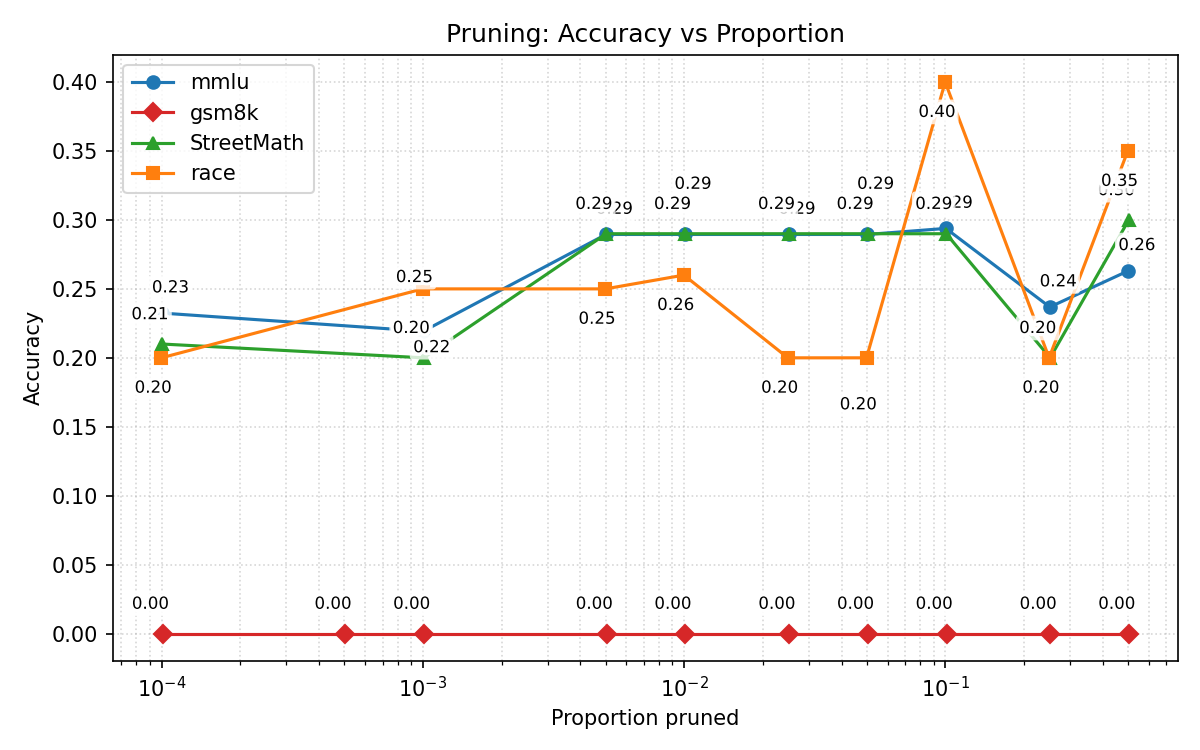}
        \subcaption{Pruning accuracy on Dream-v0-Instruct-7B}
        \label{fig:place_holder}
    \end{subfigure}

    \vspace{0.5em}

    \begin{subfigure}{0.49\textwidth}
        \centering
        \includegraphics[width=\linewidth]{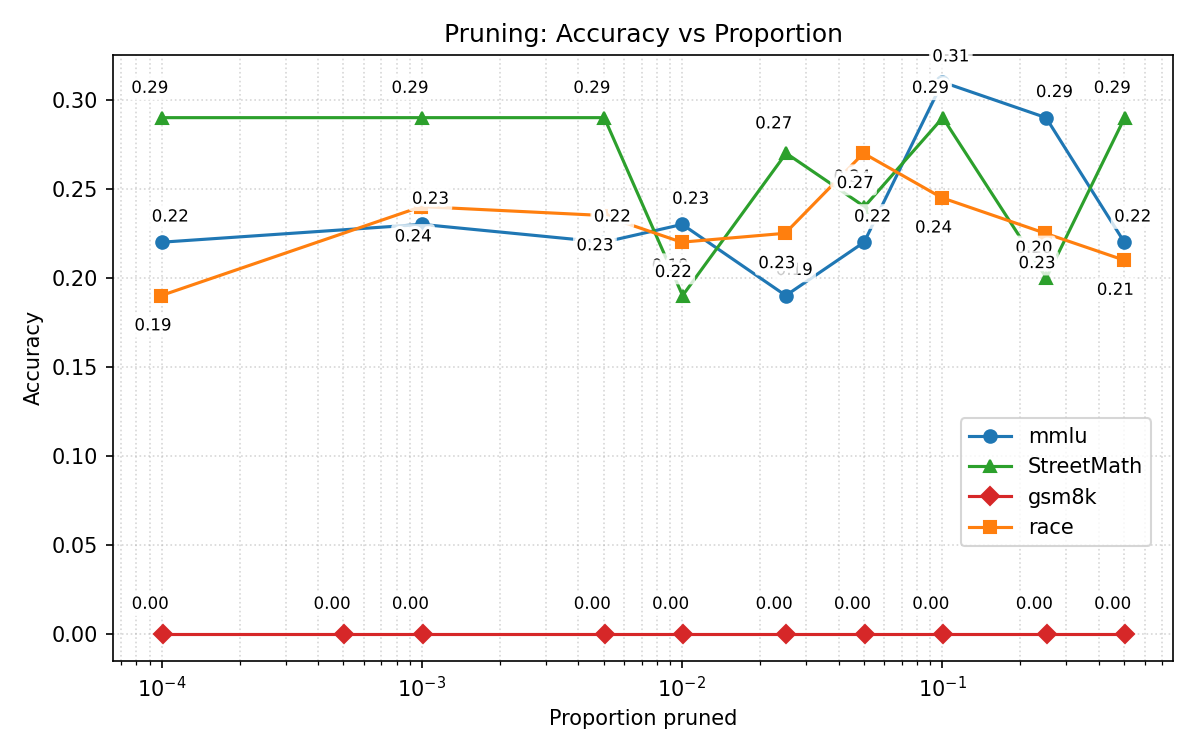}
        \subcaption{Pruning accuracy on mamba-GPT-3B.png}
        \label{fig:place_holder}
    \end{subfigure}
    \hfill
    \begin{subfigure}{0.49\textwidth}
        \centering
        \includegraphics[width=\linewidth]{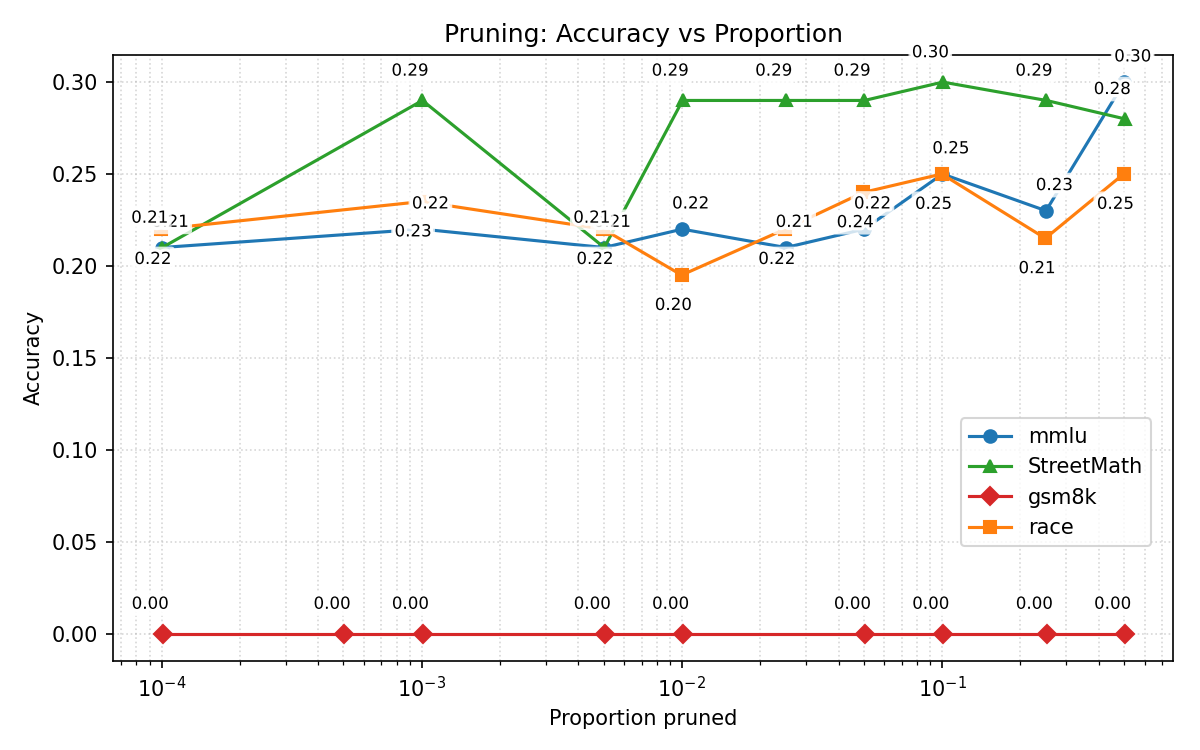}
        \subcaption{Pruning accuracy on Falcon-Mamba-7B-Instruct}
        \label{fig:place_holder}
    \end{subfigure}

    \caption{Effect of structured pruning on task performance for all models. Accuracy is plotted against the proportion of parameters pruned for StreetMath and GSM8K benchmarks.}
    \label{fig:pruning_accuracy}
\end{figure*}

We find that increasing pruning does not necessarily hurt StreetMath performance: aside from Qwen3-4B-Instruct-2507, most models remain stable or even improve under moderate pruning, contradicting the intuition that reduced capacity uniformly impairs numerical reasoning. Pruning effects diverge by benchmark, as depicted in Figure \ref{fig:pruning_accuracy}: MMLU and RACE are similarly resilient, whereas GSM8K is extremely sensitive—even slight pruning collapses accuracy to near zero across all models—implicating a specialized, fragile neuron subset for exact arithmetic while StreetMath and language-heavy tasks rely on more distributed representations. These patterns align with prior results \cite{christ_math_2025}, suggesting a dual pathway: (i) localized, brittle circuits for exact arithmetic that fail under pruning, and (ii) distributed, robust circuits for approximation and text-heavy reasoning, where moderate pruning can denoise and improve performance—consistent with StreetMath being tackled more as context-driven linguistic estimation than strict mathematical computation.

\section{Layer-wise Studies}
\begin{figure*}
\centering
 \begin{subfigure}{0.9\linewidth}
        \centering
        \includegraphics[width=\linewidth]{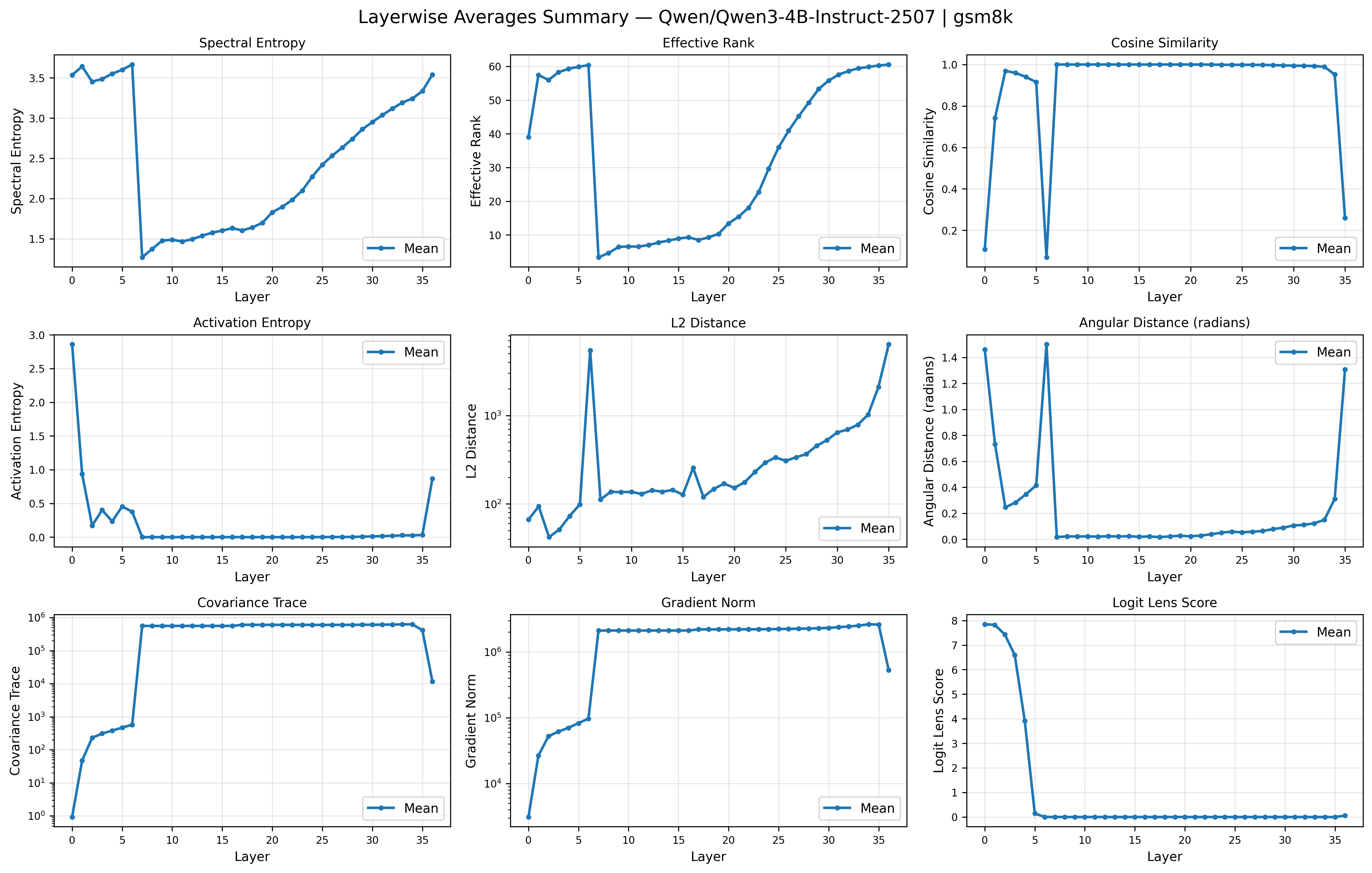}
        \subcaption{Layerwise Average Summary - Qwen3-4B-Instruct-2507 on GSM8K}
        \label{fig:place_holder}
    \end{subfigure}
    
    
    \begin{subfigure}{0.9\linewidth}
        \centering
        \includegraphics[width=\linewidth]{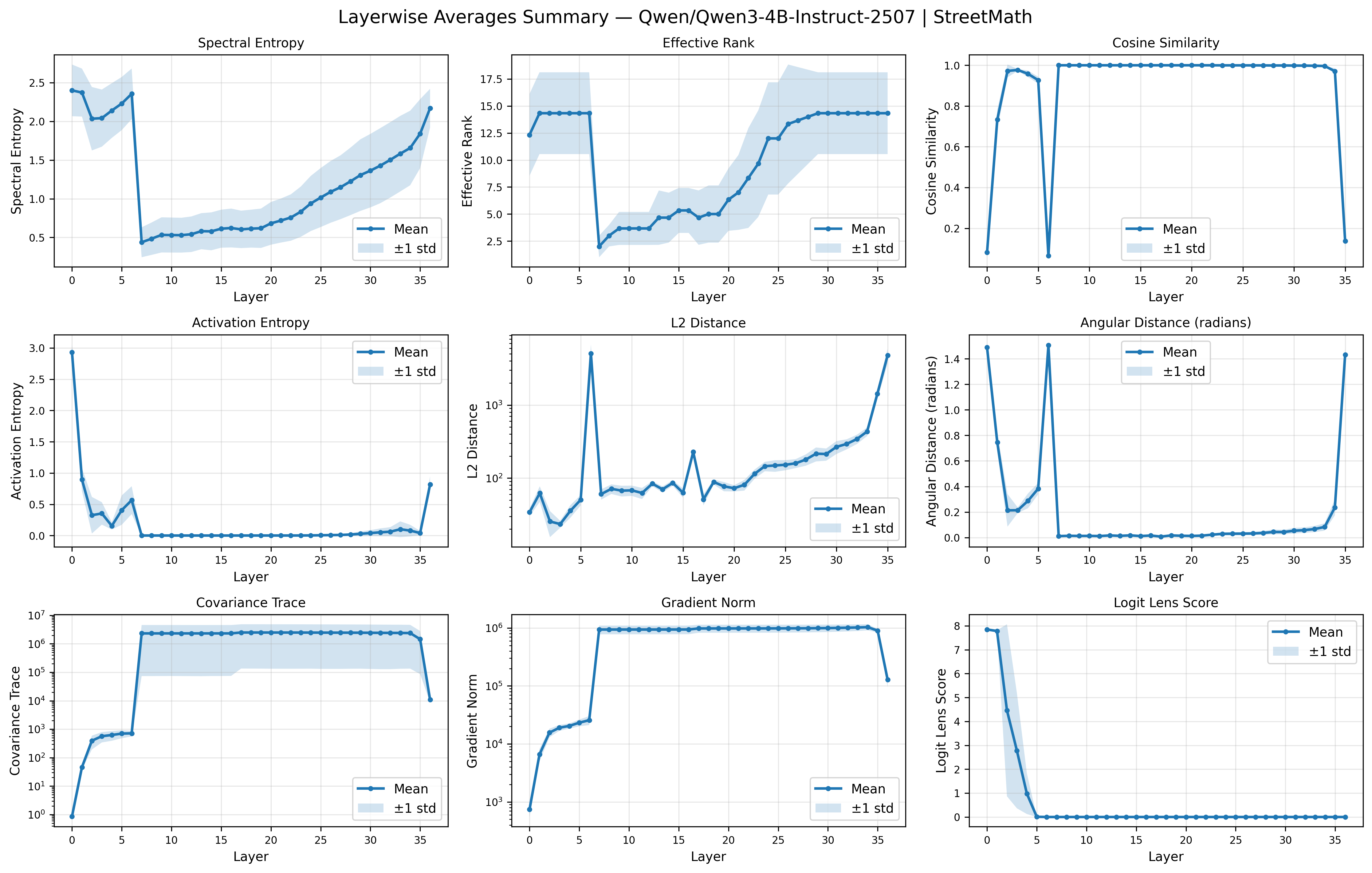}
        \subcaption{Layerwise Average Summary - Qwen3-4B-Instruct-2507 on StreetMath}
        \label{fig:place_holder}
    \end{subfigure}
    \caption{Comparative Layerwise Average Summary for Qwen3-4B-Instruct-2507 on GSM8K vs StreetMath}
    \label{fig:layerwise_studies_qwen3_instruct}
\end{figure*}

\begin{figure*}
\centering
    \begin{subfigure}{0.9\linewidth}
    \centering
    \includegraphics[width=0.9\textwidth]{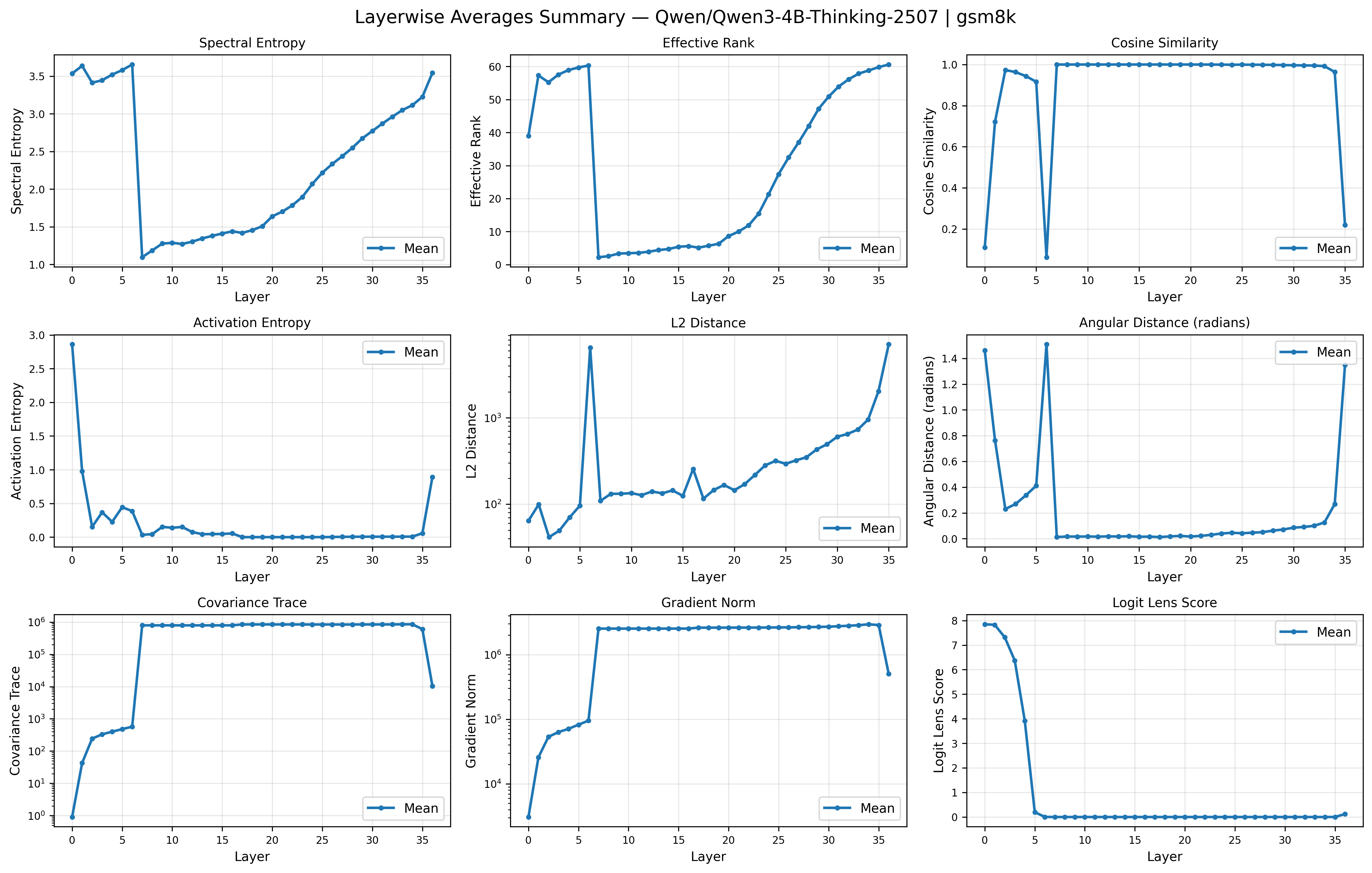}
    \caption{Layerwise Average Summary - Qwen3-4B-Thinking-2507 on GSM8K}
    \end{subfigure}

    \begin{subfigure}{0.9\linewidth}
    \centering
    \includegraphics[width=0.9\textwidth]{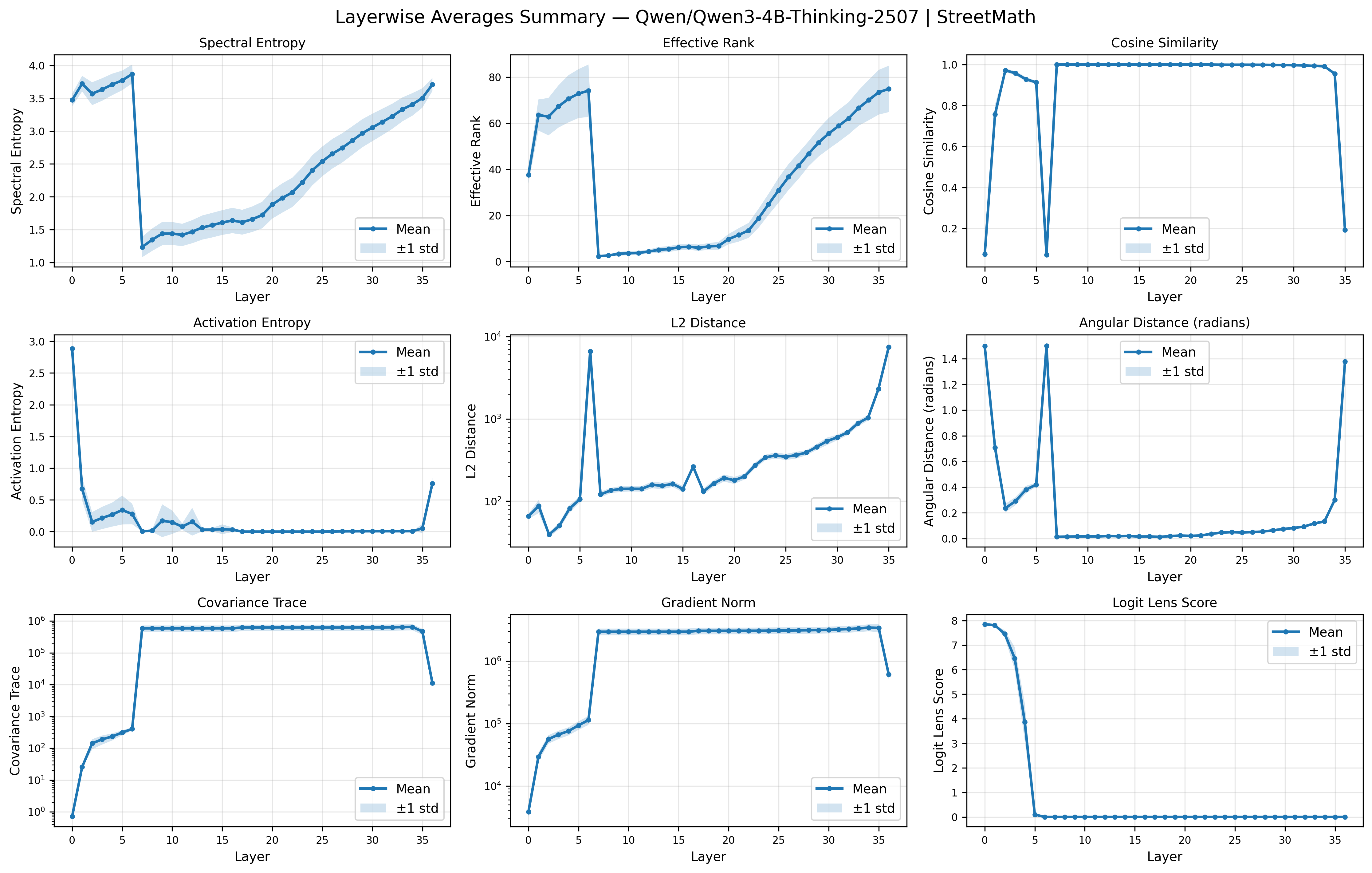}
    \caption{Layerwise Average Summary - Qwen3-4B-Thinking-2507 on StreetMath}
    \end{subfigure}

    \caption{Comparative Layerwise Average Summary for Qwen3-4B-Thinking-2507 on GSM8K vs StreetMath}
    \label{fig:layerwise_studiesqwen3_thinking}
\end{figure*}

To uncover the internal state of LLMs, we extract layerwise diagnostics from transformers on mathematical reasoning corpora and \textit{StreetMath} and analyze the spectral entropy, effective rank, activation entropy... The layer-wise analyses \cite{skean_layer_2025} reveal a broadly U-shaped evolution of spectral entropy and effective rank (high at input, dipping early, then rising) across models and tasks, with Falcon-Mamba-7B on StreetMath as the main exception, as depicted in Figure \ref{fig:layerwise_studies_qwen3_instruct} Figure \ref{fig:layerwise_studiesqwen3_thinking} and \ref{App_Layerwise}. GSM8K runs of Qwen3-4B-Instruct-2507 show a pronounced dip by the first third of layers and a steady increase. Notably, both GSM8K and StreetMath runs exhibit elbow-like transitions at comparable depths, consistent with early compression and later re-expansion seen in shortcut reasoning \cite{ding_break_2024}. This observation supports the view that approximation in StreetMath does not help models reach solutions more efficiently, showing the opposite of human cognitive miserliness~\cite{jiang_reductions_2025}.

It is evident from our experiments that task-specific effects emerge across the models. StreetMath runs typically show higher late-layer entropy and effective rank than GSM8K for the same model, along with larger transition distances. This pattern indicates not only higher variability across models but also more sustained representational expansion and stronger late-stage adjustments. By contrast, GSM8K often consolidates into a stable mid-layer corridor with very high cosine similarity and minimal angular changes. These observations support our causal study results that models use a more diverse set of neurons when handling street math-type questions while dedicating to a small set of neurons when handling exact arithmetic operations. For details, refer to \ref{App_Layerwise}.

\begin{figure*}
\centering
    \begin{subfigure}{0.9\linewidth}
    \centering
    \includegraphics[width=0.9\textwidth]{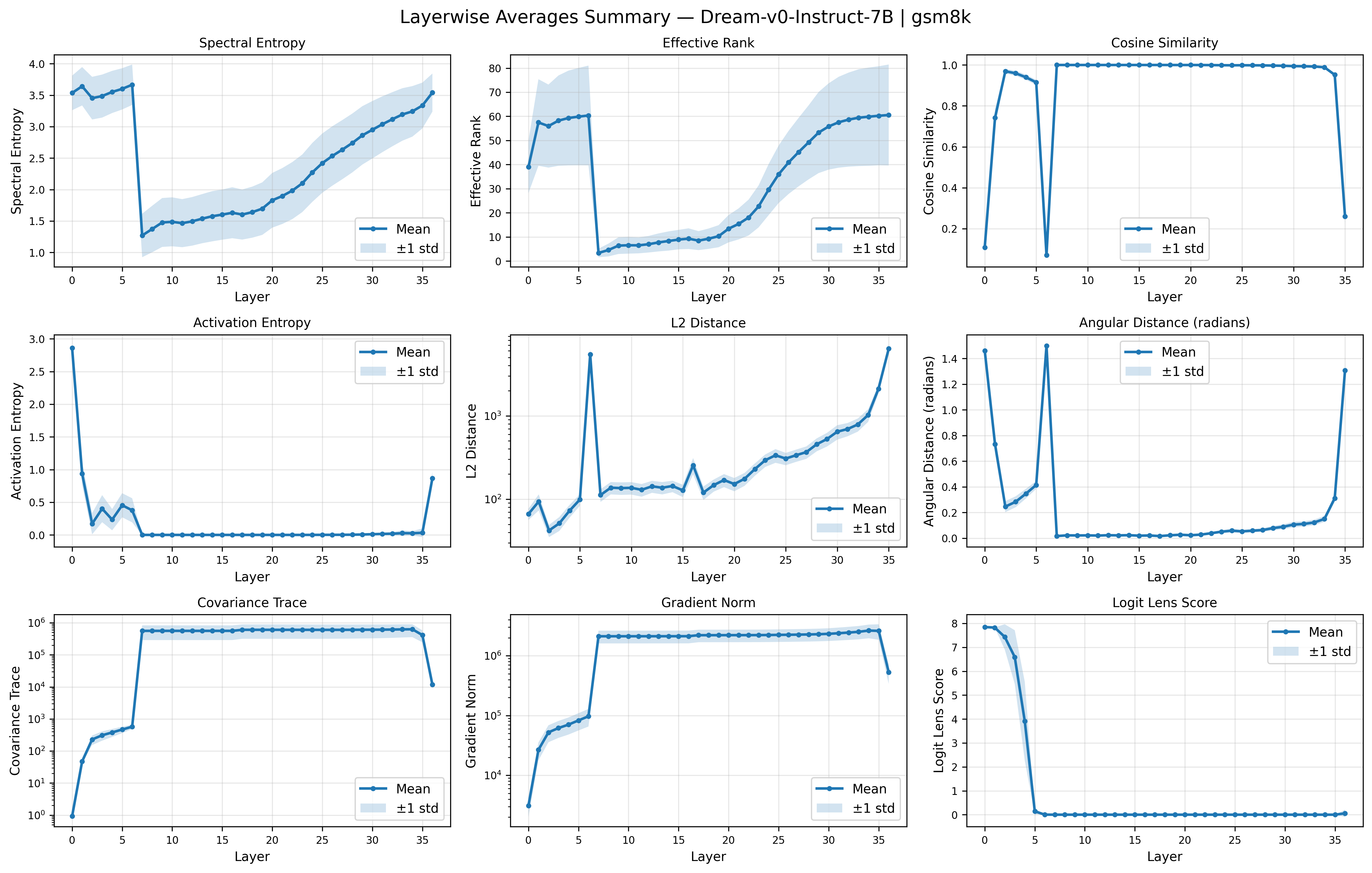}
    \caption{Layerwise Average Summary - Dream-v0-Instruct-7B on GSM8K}
    \end{subfigure}

    \begin{subfigure}{0.9\linewidth}
    \centering
    \includegraphics[width=0.9\textwidth]{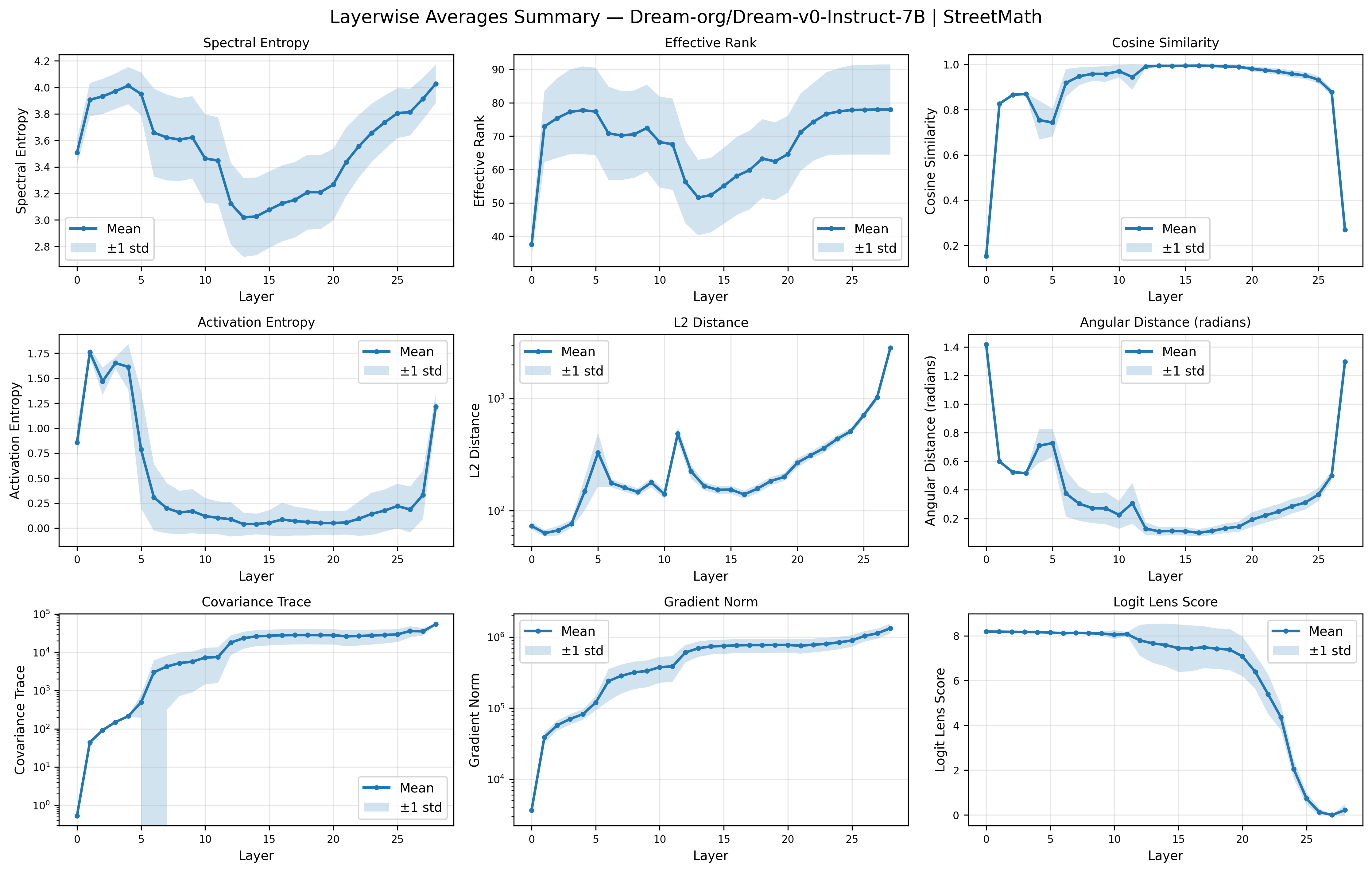}
    \caption{Layerwise Average Summary - Dream-v0-Instruct-7B on StreetMath}
    \end{subfigure}
    \caption{Comparative Layerwise Average Summary for Dream-v0-Instruct-7B on GSM8K on GSM8K vs StreetMath}
    \label{fig:layerwise_studies_dream}
\end{figure*}

\begin{figure*}
    \begin{subfigure}{\linewidth}
    \centering
    \includegraphics[width=0.9\textwidth]{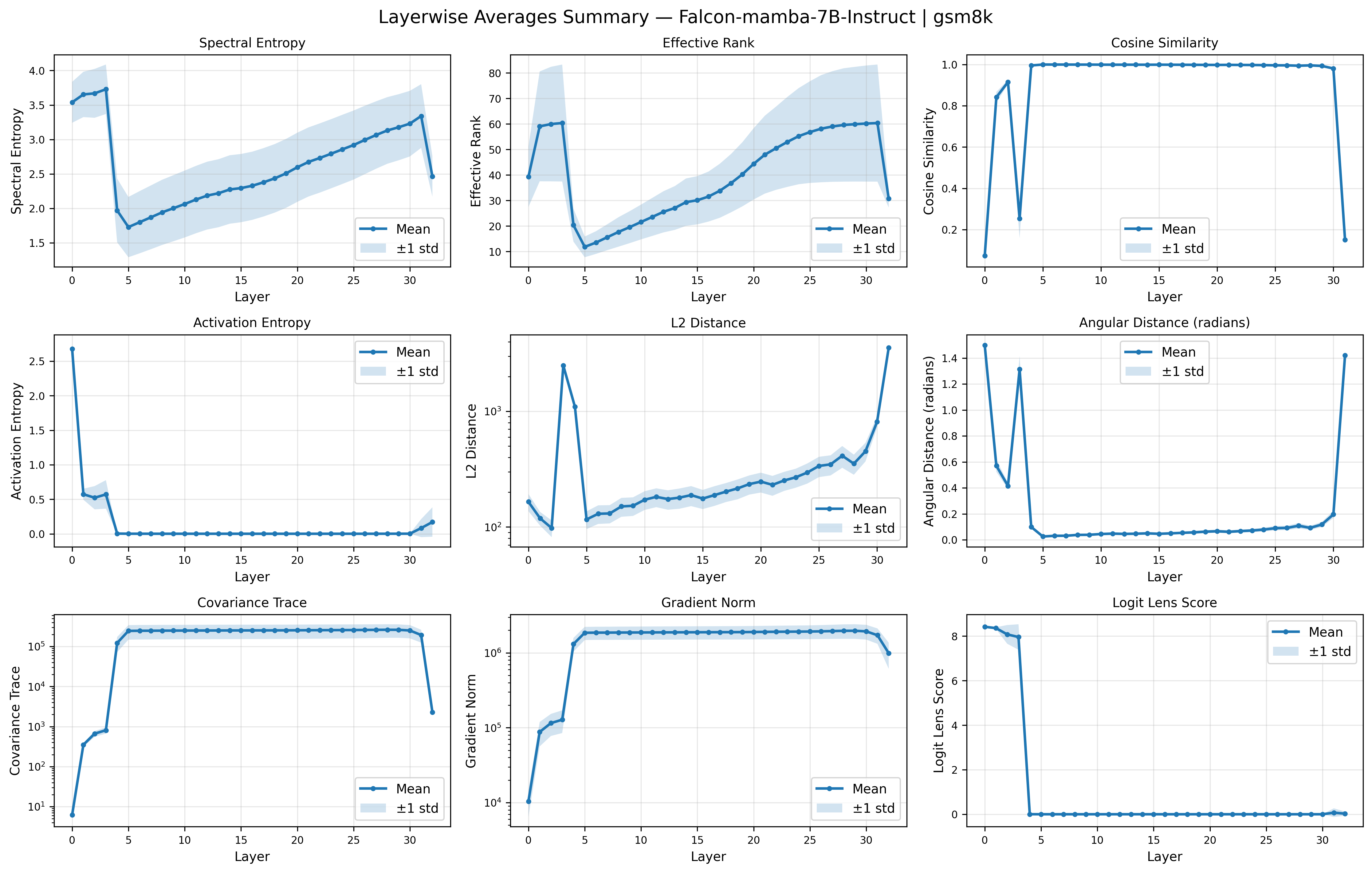}
    \caption{Layerwise Average Summary - Falcon-mamba-7B on GSM8K}
    \end{subfigure}

    \begin{subfigure}{\linewidth}
    \centering
    \includegraphics[width=0.9\textwidth]{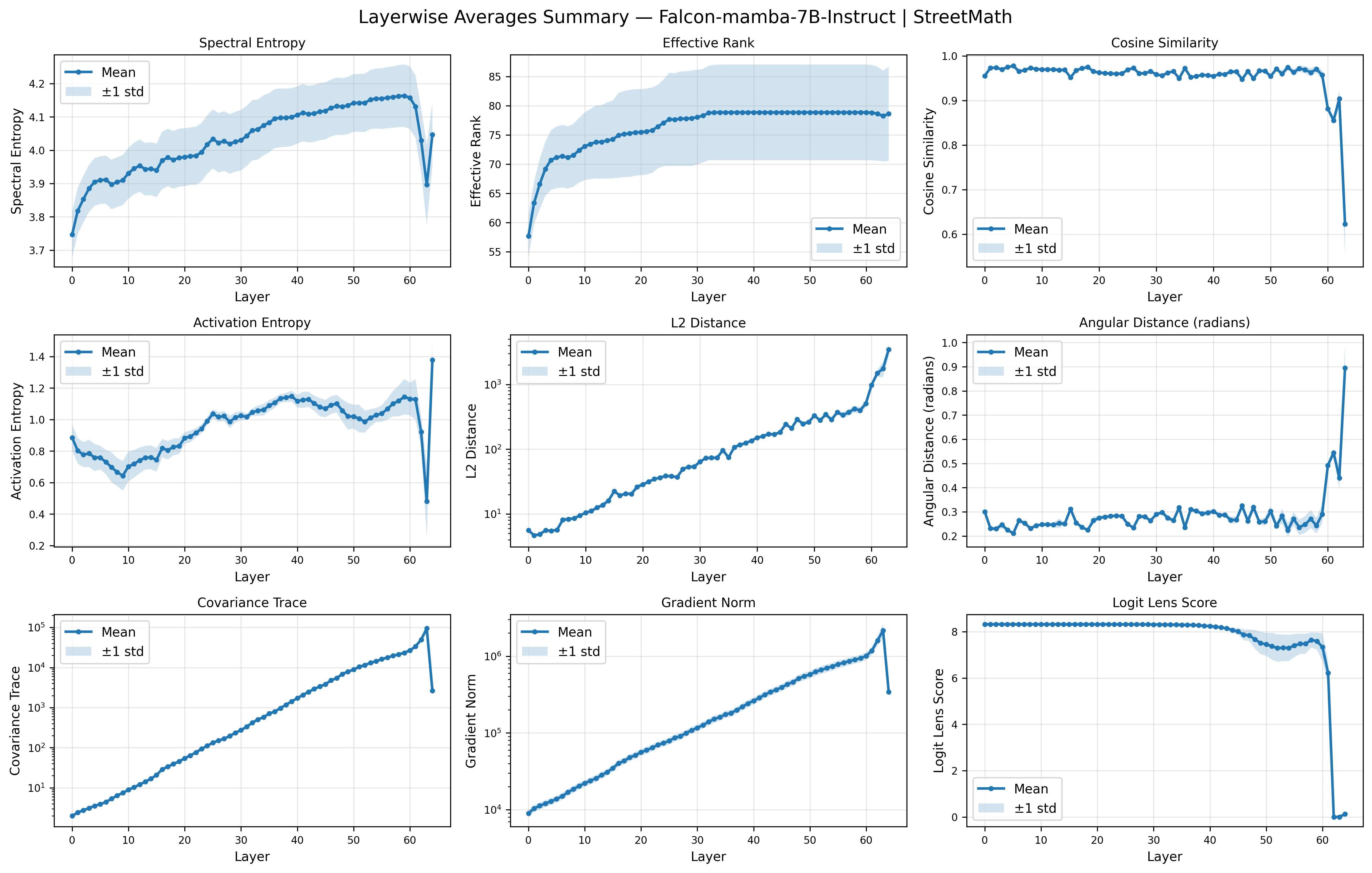}
    \caption{Layerwise Average Summary - Falcon-mamba-7B on StreetMath}
    \end{subfigure}
    \caption{Comparative Layerwise Average Summary for Falcon-mamba-7B-Instruct on GSM8K on GSM8K vs StreetMath}
    \label{fig:layerwise_studies_Falcon-mamba-7B-Instruct}
\end{figure*}

\begin{figure*}
    \begin{subfigure}{\linewidth}
    \centering
    \includegraphics[width=0.9\textwidth]{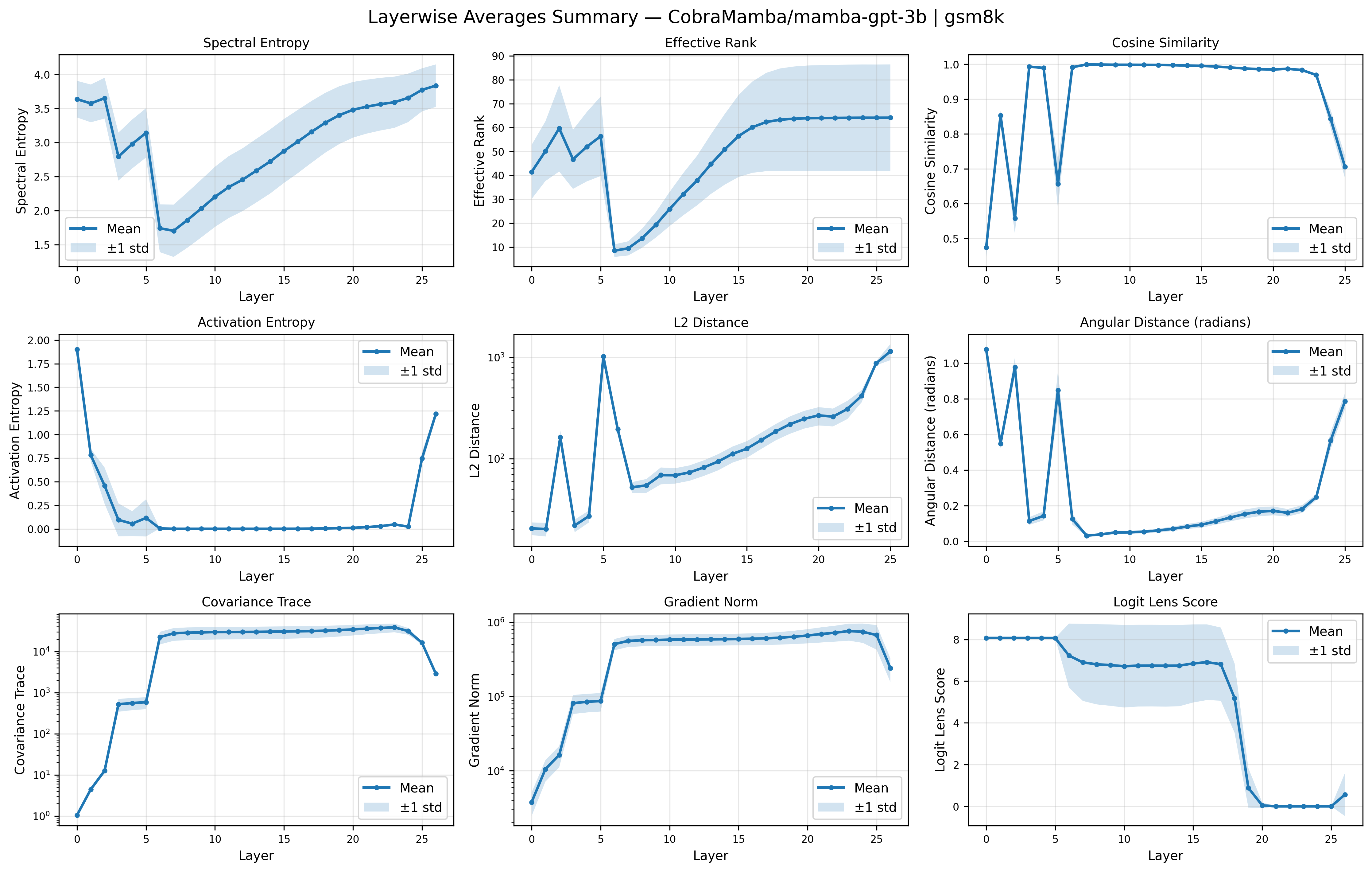}
    \caption{Layerwise Average Summary - mamba-gpt-3B on GSM8K}
    \end{subfigure}
    
    \begin{subfigure}{\linewidth}
    \centering
    \includegraphics[width=0.9\textwidth]{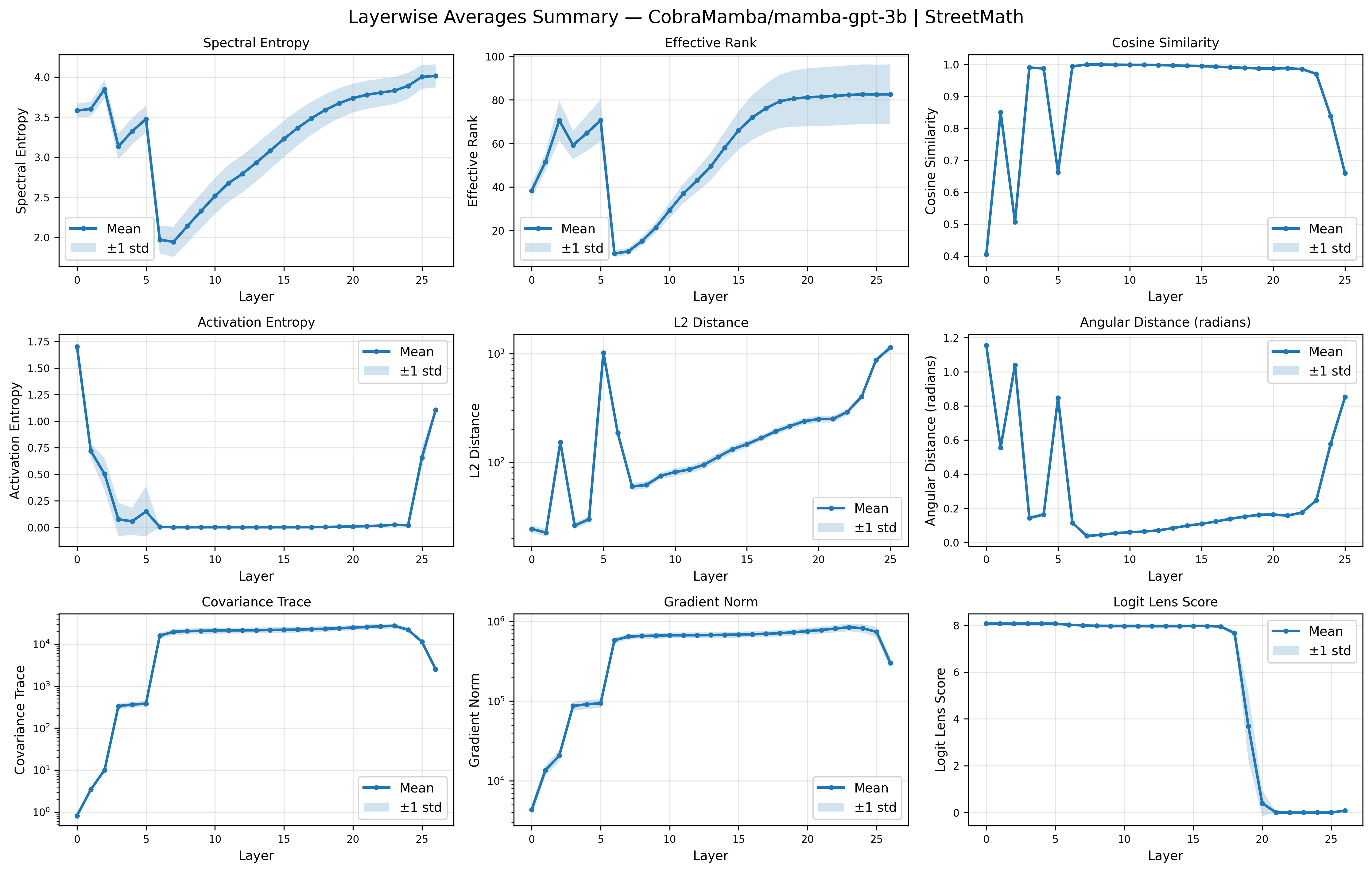}
    \caption{Layerwise Average Summary - mamba-gpt-3B on StreetMath}
    \end{subfigure}
    \caption{Comparative Layerwise Average Summary for mamba-gpt-3B on GSM8K on GSM8K vs StreetMath}
    \label{fig:layerwise_studies_mamba-gpt-3B}
\end{figure*}

\FloatBarrier

\section{Conclusion}
We curated the \textit{StreetMath} benchmark to reveal LLMs’ lack of cognitive miserliness in street-math settings. Although these models can round single numbers, they fail to use this ability to save computational effort and instead rely on exact arithmetic even when approximation would suffice. Our analyses show that models activate broader neuron sets for approximate reasoning but narrower, specialized ones for exact computation, suggesting limited flexibility in reallocating cognitive resources. Pruning experiments further indicate that removing precision-oriented parameters can improve approximation, implying that rigid numerical circuits may hinder adaptive estimation. Overall, these results demonstrate that current LLMs can perform arithmetic but not economize it—highlighting a key gap between human and machine reasoning in their ability to modulate effort based on context.

\clearpage
\appendix

\section{Experiment Setup}
\label{App_Experiment_setup}
\subsection{Model Selection}
To examine how different architectures perform under the street math setting, we selected representative models from autoregressive transformer, diffusion-based LLM, and state-space families. Given computational constraints, we restricted our study to small- and medium-sized models. To ensure reproducibility and enable deeper investigation of internal mechanisms, we further limited our selection to open-source models with publicly available weights. Because the task requires models to follow prompts reliably and generate multiple-choice responses, we focused on instruction-tuned and thinking models. Within these constraints, we also sought to preserve meaningful comparisons, such as chain-of-thought versus instruction-only models, as well as cross-architecture and cross-size contrasts.

Accordingly, our study evaluates Qwen3-4B-Instruct-2507, Qwen3-4B-Thinking-2507, Dream-v0-Instruct-7B, Falcon-Mamba-7B, and Mamba-GPT-3B. All models are initialized with the default parameters.

\subsection{Hardware specifications}
We conducted all experiments on a single NVIDIA A10 GPU hosted on RunPod, using an Ubuntu 22.04 operating system with CUDA version 12.8.1.

\section{StreetMath dataset and benchmark result}
\label{App_BenchmarkingResults}
 
\subsection{Data Curation}
StreetMath targets everyday “street math,” emphasizing fast estimation over exact arithmetic. It contains multiple-choice questions across shopping and daily-life contexts: basket totals, discounts (percentage-off, BOGO, buy-$n$-get-$m$, threshold coupons), taxes (pre/post-discount), unit conversions (lb-oz, kg-g), and tips. Prompts explicitly nudge for approximate reasoning (“about how much”) to elicit human-style rounding.

Each question has four options: the exact value; a “good approximation” within 20\% relative error (correct); a “mildly off” option; and a “way off” option (fractional or multi-fold). Choices are shuffled A–D, with metadata storing numeric values. Spacing ensures clear separation: mild $\geq 60\%$ and way $\geq 150\%$.

Good approximations follow deterministic rounding rules. Basket totals round prices to dollars, then sum and drop cents. Discounts round prices to dollars, rates to nearest 5\%, pair BOGO (buy one get one) items by price, and compute buy-$n$-get-$m$ deterministically. Threshold coupons apply to a rounded subtotal. Taxes round bases and rates (5\% steps) before dropping cents. Unit costs round prices and weights. Tips apply percentages to subtotals rounded to \$5/\$10 buckets.

Data generation is deterministic given a seed. Templates randomize prices, quantities, and rates. Outputs are JSONL lines with \texttt{id}, \texttt{topic}, \texttt{prompt}, \texttt{choices}, \texttt{labels}, \texttt{correct\_label}, and \texttt{metadata} (exact, good, mild, way). Splits are controllable by topic weights. A validator enforces spacing and alignment.

\subsection{StreetMath Benchmark}
The benchmark evaluates LLMs on StreetMath via local JSONL or hosted dataset \\(\texttt{LuxMuseAI/StreetMathDataset}). The system prompt encourages estimation and discourages exact calculation. Models must output: “Final choice: \texttt{<A|B|C|D>}”, “Answer: \texttt{<numeric>}”, and “Reasoning: \texttt{<short sentence>}”; optional inner thoughts appear in \texttt{<think>...</think>}. The runner supports OpenAI-compatible APIs, local Transformers, and Ollama.

Outputs are parsed for choice, numeric answer, reasoning, and optional tool calls. If only a number is given, the closest choice is inferred. Labels: exact = "Exact math," good = "Good approximation," mild/way = "Mildly off"/"Way off." We use the count of Good approximation as evaluation metrics to avoid giving arbitrary weights to each choice.

Each sample yields a JSON record with prompt, predictions, reasoning, token/latency, and judgement. A summary aggregates mean scores, label counts, accuracy by topic, tool-call frequency, and average resource use. This setup cleanly separates approximation skill from exact computation preference while ensuring reproducibility across models and backends.

\section{Linear Probe}
\label{App_LinearProbe}

\subsection{Experimental Setup}
\textbf{Task Definition:} We train linear probes to detect numerical proximity concepts, specifically whether numbers are "near" multiples of 5 or 10. For near-5 detection, proximity is defined as $\min(|n \bmod 10 - 0|, |n \bmod 10 - 5|, |n \bmod 10 - 10|) \leq 1$, covering digits \{0, 1, 4, 5, 6, 9\}. For near-10 detection, proximity is defined as $\min(|n \bmod 10 - 0|, |n \bmod 10 - 10|) \leq 1$, covering digits \{0, 1, 9\}.

\textbf{Data Generation:} We generated 4,000 training samples and 1,500 validation samples per condition. Numbers were randomly sampled from [0, 9999] and embedded into descriptive templates. Two template sets were used:
\begin{itemize}
\item Template A: ``Consider the number \{n\}.'', ``Let x = \{n\}.'', ``Value: \{n\}'', etc.
\item Template B: ``Here is \{n\}.'', ``We study the scalar \{n\}.'', ``Write down \{n\} and continue.'', etc.
\end{itemize}

Numbers were presented in two surface forms: digits (``25'') and words (``twenty five'') using the num2words library with normalization (hyphens and commas removed, lowercase).

\textbf{Training Protocol:} We used a two-stage streaming approach to handle memory constraints:
  \begin{enumerate}
  \item \textbf{Standardization:} StandardScaler fitted per layer using partial\_fit() with mean centering disabled
  \item \textbf{Classification:} SGD logistic regression with optimal learning rate, L2 regularization ($\alpha=10^{-4}$), and single-epoch updates
  \end{enumerate}

\subsection{Evaluation Methodology}

\textbf{Cross-Template Validation:} Three validation sets tested different robustness aspects: 1.Training: Template A + digits; 2. Validation A: Template B + digits (template robustness); 3. Validation W: Template A + words (cross-modal transfer).

\textbf{Error Analysis:} We analyzed error patterns at the best-performing layer (highest accuracy) across distance buckets. For near-5: distances 0, 1, 2+
. For near-10: distances 0-5 maintained separately. We also examined errors by rounding direction: -1 (round down closer), 0 (exact multiple), +1 (round up closer).

\textbf{Layer Selection Rationale:} We analyzed the best-performing layer rather than layer averages because: (1) it reveals models' optimal proximity detection capabilities, (2) it avoids noise from suboptimal layers that could mask genuine patterns, (3) it aligns with interpretability goals of understanding whether models \textit{can} learn proximity concepts.

\textbf{Layer Sampling:} We probed every layer (stride=1) for comprehensive analysis, skipping only embedding layers (layer 0).

\textbf{Statistical Measures:} Accuracy per layer, error rates by distance/direction, best layer identification. Results averaged over single runs with fixed random seeds (1337) for reproducibility.

\FloatBarrier

\section{Causal Study}
\label{App_Causal}
\label{CausalStudy}

Due to compute constraints, each setting is run once using bootstrap samples ($\leq 500$ examples) 
drawn from both the training set (CSV with \textit{question}, \textit{solution}, and \textit{answer} fields) 
and each calibration set. For every pruning proportion, we reload the model 
(\texttt{AutoModelForCausalLM}, \texttt{bfloat16}, \texttt{device\_map=auto}; Dream models are wrapped 
for \texttt{lm\_eval} compatibility), apply the mask, and evaluate performance using the 
\textbf{EleutherAI LM Evaluation Harness} on user-specified tasks. 

To manage compute, per-task evaluation is capped at 1{,}000 items, and prompts are truncated to 256 tokens. 
When no \texttt{lm\_eval} tasks are provided, a lightweight multiple-choice evaluator is used. 
For \textbf{GSM8K}, evaluation is limited to 1{,}000 samples. 
For \textbf{StreetMath}-style multiple choice, we treat a ``good approximation'' judgment as correct. 

All results are saved per model, per task and per pruning proportion in the specified results directory.

\section{Layerwise Study}
\label{App_Layerwise}
The experiments implement a two‑stage pipeline that first extracts layerwise diagnostics from transformer models on mathematical reasoning corpora and then aggregates and visualizes these diagnostics across many prompts.

In the first stage, model‑specific analysis scripts (for example, Dream‑v0‑Instruct‑7B, Qwen3‑4B variants, Mamba‑GPT‑3B, and Falcon‑mamba‑7B‑Instruct) load a Hugging Face model and tokenizer and evaluate it on a chosen dataset split. The workflows support both the GSM8K test split and a StreetMath test set. For each prompt, the scripts request hidden states, and compute a suite of metrics for every layer. Intra‑layer measurements include spectral entropy and effective rank ~\cite{roy2007effective} obtained from singular‑value spectra, activation entropy computed from histogram estimates, the trace of the covariance matrix as a proxy for Gaussian complexity, gradient norms approximated by the variance of hidden activations, logit‑lens proxy scores, and attention entropy when attention weights are present. Inter‑layer measurements quantify how the representation changes from one layer to the next through cosine similarity, L2 distance, and angular distance. Each prompt therefore contributes a record containing these per‑layer vectors, along with metadata, to a JSON file. Due to computational constraint, we limit each dataset to 1000 samples.

The second stage consolidates these per‑prompt records. The script reads a results JSON and computes the sample mean and the sample standard deviation across prompts for every metric and for every layer index. Because the raw results may mix series of slightly different lengths, the aggregation is performed at the most common length observed for each metric, ensuring that elementwise statistics are well‑defined and not dominated by outliers in shape. 





\section{Related Work}

\subsection{The Approximation Gap in Mathematical Reasoning}

Current mathematical reasoning research exhibits a systematic bias toward exact computation, creating a fundamental blind spot in our understanding of numerical intelligence. Zhou et al.~\cite{zhou2024pretrained} demonstrated that LLMs use specialized Fourier mechanisms for precise arithmetic, while Yu and Ananiadou~\cite{yu2024interpreting} identified localized attention heads for exact operations. Kahneman ~\cite{kahneman2011thinking}—adaptively reduces computational effort when an approximation suffices. These findings systematically overlook cognitive flexibility, instead celebrating models that can perform precise calculations while ignoring whether they can engage in the contextually appropriate approximation that characterizes genuine mathematical understanding. These mechanistic insights, while valuable, represent a narrow conception of mathematical reasoning that prioritizes precision over cognitive flexibility. Recent work by Srivastava et al. on LMThinkBench~\cite{srivastava_lmthinkbench_2024} reveals that models achieve high accuracy but at the cost of unnecessarily complex reasoning paths; a pattern consistent with systems that lack the cognitive control mechanisms necessary for adaptive approximation. When models cannot modulate their computational precision based on contextual demands, they default to maximum effort regardless of whether such precision is warranted or efficient. Highlighting the gap between computational capability and efficient reasoning. 

\subsection{Training Data Bias Toward Exact Computation}

Research reveals systematic biases in mathematical reasoning training data that favor exact computation over flexible approximation strategies. Analysis of major mathematical training corpora shows a predominant focus on problems with exact, verifiable answers. Paster et al.'s OpenWebMath dataset~\cite{paster2023openwebmath}, containing 14.7B tokens of mathematical web content, consists primarily of forum discussions, educational materials, and reference pages where mathematical problems are presented with definitive solutions rather than approximation strategies. Similarly, Lewkowycz et al.'s Minerva training corpus~\cite{lewkowycz2022minerva} drew from 118GB of scientific papers and mathematical web content that emphasizes precise computational procedures.

This training bias toward exact answers has measurable consequences for model behavior. The pattern-matching hypothesis is supported by Mirzadeh et al.'s GSM-Symbolic analysis~\cite{mirzadeh_gsm_symbolic_2024}, which reveals that model performance degrades significantly when numeric values are perturbed, indicating over-reliance on specific number patterns rather than general reasoning principles. Shao et al.~\cite{shao2024deepseekmath} explicitly acknowledge this issue, noting that their model exhibits "data selection bias in pre-training and fine-tuning" that leads to weaker performance on certain problem types.

\subsection{Overthinking and Computational Inefficiency}

Recent work has documented a troubling pattern: LLMs consistently overthink mathematical problems, generating verbose reasoning chains when simpler approaches would suffice. Ding et al.~\cite{ding2024break} proposed "break the chain" strategies to reduce token consumption, demonstrating that models maintain performance even when forced to skip intermediate steps. Zhao et al.'s work on efficiency enhancement in reasoning models~\cite{zhao_exploring_2024} suggests this isn't just a performance issue but a fundamental architectural limitation.

\subsection{Mechanistic Evidence for Competing Circuits}

Mechanistic interpretability studies reveal distinct and overlapping neural pathways for exact versus approximate reasoning. Christ et al.~\cite{christ2025math} demonstrated that math-specific parameters can be isolated through structured pruning. Skean et al.~\cite{skean2025layer} conducted a layer-by-layer analysis, revealing that different types of mathematical operations are processed at different depths in transformer architectures. Sun et al.~\cite{sun2025probing} probed arithmetic errors in language models and identified systematic patterns in computational failures, while Saynova et al.~\cite{saynova2025fact} investigated whether mathematical reasoning relies on fact recall, heuristics, or pure computation, finding evidence for multiple pathways depending on problem complexity and context.

\subsection{Numerical Representation and Geometric Understanding}

Understanding how LLMs represent numerical information has been a focus of recent mechanistic interpretability work. Levy and Geva~\cite{levy2024encode} demonstrated that language models encode numbers using individual circular representations for each digit in base 10, providing geometric understanding of numerical processing. Kantamneni and Tegmark~\cite{kantamneni2025trigon} extended this work by showing that language models use trigonometric functions in their internal computations, suggesting sophisticated geometric representations of numerical concepts. Zhu et al.~\cite{zhu2025encode} investigated how language models encode numeric magnitude, while Shah et al.~\cite{shah2023numeric} examined magnitude comparison tasks, finding that models develop specialized circuits for determining relative numerical size. These representational studies suggest that current numerical encodings may be too rigid to support flexible approximation strategies. 

\subsection{Architectural Differences in Approximation Capacity}

Different LLM architectures exhibit varying capabilities for flexible reasoning, though systematic evaluation of approximation strategies across architectures remains limited. Li et al.~\cite{li2025diffusion} explored diffusion models for language tasks, demonstrating their application to text generation, though their mathematical reasoning capabilities, particularly regarding approximation versus precision trade-offs, have not been extensively studied.

The architectural constraints that affect mathematical reasoning extend beyond approximation to fundamental information processing capabilities. Jelassi et al.~\cite{jelassi2024repeat} demonstrated that transformers can theoretically copy strings of exponential length while state-space models are fundamentally limited by their fixed-size latent state, suggesting that the rigid memory constraints that impede copying may also constrain flexible approximation strategies. These findings indicate that current architectural paradigms may systematically differ in their capacity for the kind of cognitive flexibility that characterizes human mathematical reasoning.

This architectural variation highlights a broader gap in our understanding of how different model designs affect the ability to engage in contextually appropriate approximation—a crucial aspect of mathematical intelligence that remains largely unexplored across the spectrum of current LLM architectures.

\subsection{Augmentation Strategies and Alternative Approaches}

Recognizing the limitations of pure language model approaches to arithmetic, researchers have proposed several augmentation strategies. Tool-augmented approaches represent the dominant paradigm, where models learn to invoke external calculators, symbolic solvers, or knowledge bases. Schick et al.~\cite{schick2023toolformer} introduced Toolformer, which teaches LLMs to use tools through self-supervised learning, while Das et al.~\cite{das2024mathsensei} developed MathSensei, combining web search, Python execution, and Wolfram-Alpha integration for comprehensive mathematical reasoning support.

Program-aided reasoning offers another promising direction. Gao et al.~\cite{gao2023pal} proposed Program-Aided Language models (PAL), which generate Python programs as intermediate reasoning steps, while Chen et al.~\cite{chen2022pot} introduced Program-of-Thoughts prompting to separate computation from reasoning. These approaches effectively delegate precise calculations to programming environments while preserving natural language reasoning.

At the architectural level, Dietz and Klakow~\cite{dietz2025igc} introduced the Integrated Gated Calculator (IGC), which emulates a calculator directly on the GPU, achieving 98-99\% accuracy on arithmetic tasks in a single iteration without external tools. Lauter et al.~\cite{lauter2024machine} investigated machine learning approaches for modular arithmetic, demonstrating specialized techniques for specific algebraic structures, though with limited success that highlights the inherent difficulty of certain mathematical operations.

While these augmentation strategies successfully address computational limitations and improve exact calculation capabilities, they do not resolve the fundamental issue our work identifies: the inability to engage in contextually appropriate approximation when exact computation is unnecessary. Current approaches actually reinforce the precision bias by providing increasingly sophisticated mechanisms for exact calculation, potentially exacerbating the cognitive inflexibility that characterizes current mathematical reasoning systems.

\subsection{Pattern Recognition vs. Algorithmic Understanding}

A fundamental question concerns whether models learn genuine algorithms or rely on sophisticated pattern recognition. Nikankin et al.~\cite{nikankin2025arithmetic} examined "arithmetic without algorithms," investigating whether models can perform mathematical reasoning without explicit algorithmic procedures, suggesting that models may rely on pattern recognition and approximation strategies that differ fundamentally from formal mathematical computation. Gambardella et al.~\cite{gambardella2024hard} investigated whether language models perform hard arithmetic by examining their computational processes, while Lovering et al.~\cite{lovering2024probabilities} examined language model probabilities in mathematical contexts, providing insights into how models represent uncertainty and confidence.

\subsection{The Need for Approximation-Aware Evaluation}

Current mathematical reasoning evaluation focuses exclusively on exact computation, creating a fundamental evaluation gap that obscures crucial aspects of mathematical intelligence. While Ahn et al.'s comprehensive survey~\cite{ahn_large_language_2024} emphasizes that "accuracy shouldn't be the sole metric" for evaluating mathematical reasoning and highlights the need for more robust evaluation beyond final-answer correctness, existing benchmarks continue to reward only precise answers regardless of contextual appropriateness.

This evaluation paradigm fails to assess whether LLMs can engage in the kind of flexible, context-appropriate approximation that characterizes human mathematical cognition in everyday settings. The gap is significant because it touches on fundamental questions about the nature of machine intelligence and whether current LLMs genuinely understand mathematical concepts or merely implement sophisticated pattern matching. Without evaluating approximation capabilities, we cannot determine if models possess the cognitive flexibility necessary for human-like mathematical reasoning in diverse contexts.

\section{Limitations}

While our work provides new insights into the approximation behavior of LLMs, several limitations remain. First, the \emph{StreetMath} dataset contains only 1,000 problems, which may not capture the full variety of real-world estimation tasks. Second, our evaluation focuses on a specific set of open-source models; results may not generalize to larger proprietary systems or other architectures. Third, our analysis is restricted to numerical approximation in simple arithmetic settings. Extensions to more complex mathematical domains are left for future work.

\section*{Acknowledgments}
We acknowledge the use of AI tools (ChatGPT, Codex) for text proofreading, formatting assistance and scripting.

\bibliography{StreetMath}

\begin{thebibliography}{52}
\providecommand{\natexlab}[1]{#1}
\providecommand{\url}[1]{\texttt{#1}}
\expandafter\ifx\csname urlstyle\endcsname\relax
  \providecommand{\doi}[1]{doi: #1}\else
  \providecommand{\doi}{doi: \begingroup \urlstyle{rm}\Url}\fi

\bibitem[Ahn et~al.(2024)Ahn, Verma, Lou, Liu, Zhang, and Yin]{ahn_large_language_2024}
Janice Ahn, Rishu Verma, Renze Lou, Di~Liu, Rui Zhang, and Wenpeng Yin.
\newblock Large language models for mathematical reasoning: Progresses and challenges, February 2024.
\newblock URL \url{http://arxiv.org/abs/2402.00157}.
\newblock arXiv:2402.00157.

\bibitem[Alain and Bengio(2016)]{alain2016understanding}
Guillaume Alain and Yoshua Bengio.
\newblock Understanding intermediate layers using linear classifier probes.
\newblock \emph{arXiv preprint arXiv:1610.01644}, 2016.

\bibitem[Belinkov and Glass(2019)]{belinkov2019analysis}
Yonatan Belinkov and James Glass.
\newblock Analysis methods in neural language processing: A survey.
\newblock \emph{Transactions of the Association for Computational Linguistics}, 7:\penalty0 49--72, 2019.

\bibitem[Chen et~al.(2022)Chen, Ma, Wang, and Cohen]{chen2022pot}
Wenhu Chen, Xueguang Ma, Xinyi Wang, and William~W Cohen.
\newblock Program of thoughts prompting: Disentangling computation from reasoning for numerical reasoning tasks.
\newblock \emph{arXiv preprint arXiv:2211.12588}, 2022.

\bibitem[Christ et~al.(2025{\natexlab{a}})Christ, Gottesman, Kropko, and Hartvigsen]{christ2025math}
B.~R. Christ, Z.~Gottesman, J.~Kropko, and T.~Hartvigsen.
\newblock Math neurosurgery: Isolating language models' math reasoning abilities using only forward passes.
\newblock \emph{arXiv preprint}, 2025{\natexlab{a}}.

\bibitem[Christ et~al.(2025{\natexlab{b}})Christ, Gottesman, Kropko, and Hartvigsen]{christ_math_2025}
Bryan~R. Christ, Zack Gottesman, Jonathan Kropko, and Thomas Hartvigsen.
\newblock Math neurosurgery: Isolating language models' math reasoning abilities using only forward passes, June 2025{\natexlab{b}}.
\newblock URL \url{http://arxiv.org/abs/2410.16930}.
\newblock arXiv:2410.16930.

\bibitem[CobraMamba(2023)]{mambagpt3b_huggingface}
CobraMamba.
\newblock Mamba-gpt-3b, 2023.
\newblock Hugging Face model card; Apache-2.0 license.

\bibitem[Das et~al.(2024)Das, Banerjee, Manocha, and Baral]{das2024mathsensei}
Debrup Das, Debopriyo Banerjee, Somak Manocha, and Ashish Baral.
\newblock Mathsensei: A tool-augmented large language model for mathematical reasoning.
\newblock \emph{arXiv preprint arXiv:2402.17231}, 2024.

\bibitem[De~Brauwer et~al.(2006)De~Brauwer, Verguts, and Fias]{debrauwer2006five}
Johan De~Brauwer, Tom Verguts, and Wim Fias.
\newblock The representation of multiplication facts: Developmental changes in the problem size, five, and tie effects.
\newblock \emph{Journal of Experimental Child Psychology}, 94\penalty0 (1):\penalty0 43--66, 2006.

\bibitem[Dehaene(2011)]{dehaene2011number}
Stanislas Dehaene.
\newblock \emph{The number sense: How the mind creates mathematics}.
\newblock OUP USA, 2011.

\bibitem[Dietz and Klakow(2025)]{dietz2025igc}
M.~Dietz and D.~Klakow.
\newblock Igc: Integrating a gated calculator.
\newblock \emph{arXiv preprint}, 2025.

\bibitem[Ding et~al.(2024{\natexlab{a}})Ding, Liu, Fu, Song, Xie, and Zhang]{ding_break_2024}
Mengru Ding, Hanmeng Liu, Zhizhang Fu, Jian Song, Wenbo Xie, and Yue Zhang.
\newblock Break the chain: Large language models can be shortcut reasoners, June 2024{\natexlab{a}}.
\newblock URL \url{http://arxiv.org/abs/2406.06580}.
\newblock arXiv:2406.06580.

\bibitem[Ding et~al.(2024{\natexlab{b}})]{ding2024break}
Y.~Ding et~al.
\newblock Break the chain: Large language models with heuristics.
\newblock \emph{arXiv preprint}, 2024{\natexlab{b}}.

\bibitem[Fiske and Taylor(1991)]{fiske1991social}
Susan~T. Fiske and Shelley~E. Taylor.
\newblock \emph{Social Cognition}.
\newblock McGraw-Hill Series in Social Psychology. McGraw-Hill, New York, 2nd edition, 1991.

\bibitem[Gambardella et~al.(2024)Gambardella, Iwasawa, and Matsuo]{gambardella2024hard}
Andrew Gambardella, Yusuke Iwasawa, and Yutaka Matsuo.
\newblock Language models do hard arithmetic tasks easily and hardly do easy arithmetic tasks.
\newblock In \emph{Proceedings of the 62nd Annual Meeting of the Association for Computational Linguistics (Volume 2: Short Papers)}, pages 811--824, Bangkok, Thailand, 2024. Association for Computational Linguistics.
\newblock \doi{10.18653/v1/2024.acl-short.74}.
\newblock URL \url{https://aclanthology.org/2024.acl-short.74/}.

\bibitem[Gao et~al.(2023)Gao, Madaan, Zhou, Alon, Liu, Yang, Callan, and Neubig]{gao2023pal}
Luyu Gao, Aman Madaan, Shuyan Zhou, Uri Alon, Pengfei Liu, Yiming Yang, Jamie Callan, and Graham Neubig.
\newblock Pal: Program-aided language models.
\newblock In \emph{Proceedings of the 40th International Conference on Machine Learning}, volume 202 of \emph{Proceedings of Machine Learning Research}, pages 10764--10779. PMLR, 2023.
\newblock URL \url{https://proceedings.mlr.press/v202/gao23f.html}.

\bibitem[Goldberg(2016)]{goldberg2016primer}
Yoav Goldberg.
\newblock A primer on neural network models for natural language processing.
\newblock \emph{Journal of Artificial Intelligence Research}, 57:\penalty0 345--420, 2016.

\bibitem[Hewitt and Manning(2019)]{hewitt2019structural}
John Hewitt and Christopher~D Manning.
\newblock A structural probe for finding syntax in word representations.
\newblock In \emph{Proceedings of the 2019 Conference of the North American Chapter of the Association for Computational Linguistics: Human Language Technologies, Volume 1 (Long and Short Papers)}, pages 4129--4138, 2019.

\bibitem[Jelassi et~al.(2024)Jelassi, Brandfonbrener, Kakade, and Malach]{jelassi2024repeat}
Samy Jelassi, David Brandfonbrener, Sham~M. Kakade, and Eran Malach.
\newblock Repeat after me: Transformers are better than state space models at copying.
\newblock In \emph{Proceedings of the 41st International Conference on Machine Learning}, volume 235 of \emph{Proceedings of Machine Learning Research}, pages 21502--21521. PMLR, 2024.
\newblock URL \url{https://proceedings.mlr.press/v235/jelassi24a.html}.

\bibitem[Jiang et~al.(2025)Jiang, Ye, Zhao, and Gu]{jiang_reductions_2025}
Dorothy~Lianlian Jiang, Shun Ye, Liang Zhao, and Bin Gu.
\newblock Do reductions in search costs for partial information on online platforms lead to better consumer decisions? evidence of cognitive miser behavior from a natural experiment.
\newblock page isre.2022.0432, February 2025.
\newblock ISSN 1047-7047, 1526-5536.
\newblock \doi{10.1287/isre.2022.0432}.
\newblock URL \url{https://pubsonline.informs.org/doi/10.1287/isre.2022.0432}.

\bibitem[Kahneman(2011)]{kahneman2011thinking}
Daniel Kahneman.
\newblock \emph{Thinking, fast and slow}.
\newblock Farrar, Straus and Giroux, 2011.

\bibitem[Kantamneni and Tegmark(2025)]{kantamneni2025trigon}
S.~Kantamneni and Max Tegmark.
\newblock Language models use trigonometric functions.
\newblock \emph{arXiv preprint}, 2025.

\bibitem[Lauter et~al.(2024)]{lauter2024machine}
K.~Lauter et~al.
\newblock Machine learning for modular arithmetic.
\newblock \emph{arXiv preprint}, 2024.

\bibitem[Levy and Geva(2025)]{levy2025language}
Amit~Arnold Levy and Mor Geva.
\newblock Language models encode numbers using digit representations in base 10.
\newblock In \emph{Proceedings of the 2025 Conference of the Nations of the Americas Chapter of the Association for Computational Linguistics: Human Language Technologies (Volume 2: Short Papers)}, pages 385--395, 2025.

\bibitem[Levy and Geva(2024)]{levy2024encode}
Omer Levy and Mor Geva.
\newblock Language models encode numbers.
\newblock \emph{arXiv preprint}, 2024.

\bibitem[Lewkowycz et~al.(2022)Lewkowycz, Andreassen, Dohan, Dyer, Michalewski, Ramasesh, Slone, Anil, Schlag, Gutman-Solo, et~al.]{lewkowycz2022minerva}
Aitor Lewkowycz, Anders Andreassen, David Dohan, Ethan Dyer, Henryk Michalewski, Vinay Ramasesh, Ambrose Slone, Cem Anil, Imanol Schlag, Theo Gutman-Solo, et~al.
\newblock Solving quantitative reasoning problems with language models.
\newblock \emph{Advances in Neural Information Processing Systems}, 35:\penalty0 3843--3857, 2022.

\bibitem[Li et~al.(2025)]{li2025diffusion}
J.~Li et~al.
\newblock Diffusion language models.
\newblock \emph{arXiv preprint}, 2025.

\bibitem[Lovering et~al.(2024{\natexlab{a}})]{lovering2024probabilities}
C.~Lovering et~al.
\newblock Language model probabilities.
\newblock \emph{arXiv preprint}, 2024{\natexlab{a}}.

\bibitem[Lovering et~al.(2024{\natexlab{b}})Lovering, Krumdick, Lai, Ebner, Kumar, Reddy, Koncel-Kedziorski, and Tanner]{lovering2024language}
Charles Lovering, Michael Krumdick, Viet~Dac Lai, Seth Ebner, Nilesh Kumar, Varshini Reddy, Rik Koncel-Kedziorski, and Chris Tanner.
\newblock Language model probabilities are not calibrated in numeric contexts.
\newblock \emph{arXiv preprint arXiv:2410.16007}, 2024{\natexlab{b}}.

\bibitem[McCoy et~al.(2019)McCoy, Pavlick, and Linzen]{mccoy2019right}
R~Thomas McCoy, Ellie Pavlick, and Tal Linzen.
\newblock Right for the wrong reasons: Diagnosing syntactic heuristics in natural language inference.
\newblock \emph{arXiv preprint arXiv:1902.01007}, 2019.

\bibitem[Mirzadeh et~al.(2024)Mirzadeh, Alizadeh, Shahrokhi, Tuzel, Bengio, and Farajtabar]{mirzadeh_gsm_symbolic_2024}
Iman Mirzadeh, Keivan Alizadeh, Hooman Shahrokhi, Oncel Tuzel, Samy Bengio, and Mehrdad Farajtabar.
\newblock {GSM-Symbolic}: Understanding the limitations of mathematical reasoning in large language models, October 2024.
\newblock URL \url{http://arxiv.org/abs/2410.05229}.
\newblock Apple; arXiv:2410.05229.

\bibitem[Moyer and Landauer(1967)]{moyer1967time}
Robert~S Moyer and Thomas~K Landauer.
\newblock Time required for judgements of numerical inequality.
\newblock \emph{Nature}, 215\penalty0 (5109):\penalty0 1519--1520, 1967.

\bibitem[Nikankin et~al.(2025)]{nikankin2025arithmetic}
A.~Nikankin et~al.
\newblock Arithmetic without algorithms.
\newblock \emph{arXiv preprint}, 2025.

\bibitem[Paster et~al.(2023)Paster, Santos, Azerbayev, and Ba]{paster2023openwebmath}
Keiran Paster, Marco~Dos Santos, Zhangir Azerbayev, and Jimmy Ba.
\newblock Openwebmath: An open dataset of high-quality mathematical web text.
\newblock \emph{arXiv preprint arXiv:2310.06786}, 2023.

\bibitem[Rai et~al.(2025)Rai, Zhou, Feng, Saparov, and Yao]{rai_practical_2025}
Daking Rai, Yilun Zhou, Shi Feng, Abulhair Saparov, and Ziyu Yao.
\newblock A practical review of mechanistic interpretability for transformer-based language models, March 2025.
\newblock URL \url{http://arxiv.org/abs/2407.02646}.
\newblock arXiv:2407.02646.

\bibitem[Roy and Vetterli(2007)]{roy2007effective}
O.~Roy and M.~Vetterli.
\newblock The effective rank: A measure of effective dimensionality.
\newblock In \emph{2007 15th European Signal Processing Conference}, pages 606--610. IEEE, 2007.

\bibitem[Saynova et~al.(2025)]{saynova2025fact}
A.~Saynova et~al.
\newblock Fact recall, heuristics or pure computation.
\newblock \emph{arXiv preprint}, 2025.

\bibitem[Schick et~al.(2023)Schick, Dwivedi-Yu, Dess{\`a}, Raileanu, Lomeli, Zettlemoyer, Cancedda, and Scialom]{schick2023toolformer}
Timo Schick, Jane Dwivedi-Yu, Roberto Dess{\`a}, Roberta Raileanu, Maria Lomeli, Luke Zettlemoyer, Nicola Cancedda, and Thomas Scialom.
\newblock Toolformer: Language models can teach themselves to use tools.
\newblock \emph{arXiv preprint arXiv:2302.04761}, 2023.

\bibitem[Shah et~al.(2023)Shah, Marupudi, Koenen, Bhardwaj, and Varma]{shah2023numeric}
Raj Shah, Vijay Marupudi, Reba Koenen, Khushi Bhardwaj, and Sashank Varma.
\newblock Numeric magnitude comparison effects in large language models.
\newblock In \emph{Findings of the Association for Computational Linguistics: ACL 2023}, pages 6147--6161, Toronto, Canada, 2023. Association for Computational Linguistics.
\newblock \doi{10.18653/v1/2023.findings-acl.383}.
\newblock URL \url{https://aclanthology.org/2023.findings-acl.383/}.

\bibitem[Shao et~al.(2024)Shao, Wang, Zhu, Xu, Song, Zhang, Li, Gong, Jin, Wang, et~al.]{shao2024deepseekmath}
Zhihong Shao, Peiyi Wang, Qihao Zhu, Runxin Xu, Junxiao Song, Mingchuan Zhang, Y.~K. Li, Yihan Gong, Zihan Jin, Xiao Wang, et~al.
\newblock Deepseekmath: Pushing the limits of mathematical reasoning in open language models.
\newblock \emph{arXiv preprint arXiv:2402.03300}, 2024.

\bibitem[Skean et~al.(2025{\natexlab{a}})]{skean2025layer}
M.~Skean et~al.
\newblock Layer by layer: Uncovering mathematical reasoning.
\newblock \emph{arXiv preprint}, 2025{\natexlab{a}}.

\bibitem[Skean et~al.(2025{\natexlab{b}})Skean, Arefin, Zhao, Patel, Naghiyev, {LeCun}, and Shwartz-Ziv]{skean_layer_2025}
Oscar Skean, Md~Rifat Arefin, Dan Zhao, Niket Patel, Jalal Naghiyev, Yann {LeCun}, and Ravid Shwartz-Ziv.
\newblock Layer by layer: Uncovering hidden representations in language models, June 2025{\natexlab{b}}.
\newblock URL \url{http://arxiv.org/abs/2502.02013}.
\newblock version: 2; arXiv:2502.02013.

\bibitem[Srivastava et~al.(2024)Srivastava, Hussain, Srinivasan, and Wang]{srivastava_lmthinkbench_2024}
Gaurav Srivastava, Aafiya Hussain, Sriram Srinivasan, and Xuan Wang.
\newblock {LMThinkBench}: Towards basic math reasoning and overthinking in large language models, July 2024.
\newblock URL \url{http://arxiv.org/abs/2507.04023}.
\newblock arXiv:2507.04023.

\bibitem[Sun et~al.(2025)]{sun2025probing}
X.~Sun et~al.
\newblock Probing for arithmetic errors in language models.
\newblock \emph{arXiv preprint}, 2025.

\bibitem[Team(2025)]{qwen3technicalreport}
Qwen Team.
\newblock Qwen3 technical report, 2025.
\newblock URL \url{https://arxiv.org/abs/2505.09388}.
\newblock arXiv:2505.09388.

\bibitem[Teerapittayanon et~al.(2016)Teerapittayanon, McDanel, and Kung]{teerapittayanon2016branchynet}
Surat Teerapittayanon, Bradley McDanel, and Hsiang-Tsung Kung.
\newblock Branchynet: Fast inference via early exiting from deep neural networks.
\newblock In \emph{2016 23rd international conference on pattern recognition (ICPR)}, pages 2464--2469. IEEE, 2016.

\bibitem[Ye et~al.(2025)Ye, Xie, Zheng, Gao, Wu, Jiang, Li, and Kong]{ye2025dream}
Jiacheng Ye, Zhihui Xie, Lin Zheng, Jiahui Gao, Zirui Wu, Xin Jiang, Zhenguo Li, and Lingpeng Kong.
\newblock Dream 7b: Diffusion large language models.
\newblock \emph{arXiv preprint arXiv:2508.15487}, 2025.

\bibitem[Yu and Ananiadou(2024)]{yu2024interpreting}
Zeping Yu and Sophia Ananiadou.
\newblock Interpreting arithmetic mechanism in large language models through comparative neuron analysis.
\newblock In \emph{Proceedings of the 2024 Conference on Empirical Methods in Natural Language Processing}, pages 3293--3306, 2024.

\bibitem[Zhao et~al.(2024)Zhao, Guo, Deng, Sui, Hu, Zhao, Che, Qin, Chua, and Liu]{zhao_exploring_2024}
Weixiang Zhao, Jiahe Guo, Yang Deng, Xingyu Sui, Yulin Hu, Yanyan Zhao, Wanxiang Che, Bing Qin, Tat-Seng Chua, and Ting Liu.
\newblock Exploring and exploiting the inherent efficiency within large reasoning models for self-guided efficiency enhancement, June 2024.
\newblock URL \url{http://arxiv.org/abs/2506.15647}.
\newblock arXiv:2506.15647.

\bibitem[Zhou et~al.(2024)Zhou, Fu, Sharan, and Jia]{zhou2024pretrained}
Tianyi Zhou, Deqing Fu, Vatsal Sharan, and Robin Jia.
\newblock Pre-trained large language models use fourier features to compute addition.
\newblock \emph{arXiv preprint arXiv:2406.03445}, 2024.

\bibitem[Zhu et~al.(2025)]{zhu2025encode}
W.~Zhu et~al.
\newblock Language models encode the concept of numeric magnitude.
\newblock \emph{arXiv preprint}, 2025.

\bibitem[Zuo et~al.(2024)Zuo, Velikanov, Rhaiem, Chahed, Belkada, Kunsch, and Hacid]{zuo2024falconmambacompetitiveattentionfree}
Jingwei Zuo, Maksim Velikanov, Dhia~Eddine Rhaiem, Ilyas Chahed, Younes Belkada, Guillaume Kunsch, and Hakim Hacid.
\newblock Falcon mamba: The first competitive attention-free 7b language model, 2024.
\newblock URL \url{https://arxiv.org/abs/2410.05355}.
\newblock arXiv:2410.05355.

\end{thebibliography}

\end{document}